# Intercomparison of Machine Learning Algorithms for Remote Sensing-based In-season Crop Mapping

August Posch, Jitendra Kumar, Forrest M. Hoffman, Auroop R. Ganguly

***Abstract*—In-season crop type mapping is critical for food security in the face of increasingly extreme climate-related threats to crops. Currently, the USDA Cropland Data Layer provides crop type labels at 30m resolution and is available the February after harvest, but no product exists that maps crop types before harvest with satisfactory accuracy that would allow emergency managers to respond to crop threats in near real time. Furthermore, the relative advantages of a wide range of algorithms have not been evaluated in a way that accounts for interannual variability, until this study. Here, Harmonized Landsat-Sentinel surface reflectance imagery time series and crop rotation history information are combined to map corn in Iowa and almonds in California at 30m resolution accurately by early June in unseen years, with robust quantification of uncertainty due to phenology and crop distribution. Thousands of model configurations across ten machine learning algorithms were compared using a year-wise cross-validation and a suite of metrics. Hyperparameter search revealed Support Vector Machines to be the most successful algorithm overall, with a mean F1 score of 0.74 (0.59) across five unseen validation years for almonds by early June in California (corn by early June in Iowa). Interannual variation was a large source of uncertainty, but patterns showed the potential to further improve performance with ensemble approaches or ancillary data. Future work may extend these methods to include multiclass maps of all crop types, CONUS-wide maps, and in-season crop yield forecasting.**

## I. INTRODUCTION

Rapidly changing weather and climate conditions wreak havoc on natural resources in the US by increasing the risk of food shortages, economic collapse, and infrastructure damage. Extreme riverine floods [1] and dangerously shifting insect populations [2] due to climate change may wipe out specific crops and farmland regions [3], leading to food shortages, higher prices for consumers, and a loss of agriculture sector jobs. Without adaptation, climate change may reduce yields by 7-23%, and machine learning approaches can be integrated with high-resolution crop phenology information into crop models to mitigate this yield reduction [4]. The USDA publishes the Cropland Data Layer (CDL), a publicly available tool for researchers, farmers, and policymakers to assess spatial crop distribution across the conterminous United States (CONUS), but it is available only after the end of the season, typically the February after harvest [5]. Responding to natural hazards as they happen requires an early and in-season map instead. Thus, to preserve livelihoods in the agriculture sector and ensure consumer price stability, an in-season crop map is necessary to assess impacts of extreme weather, forecast yields, and inform food security decisions.

Current state of the art studies have proposed machine learning methods to create in-season maps, using an end-of-season product like the CDL as a reference labeled dataset, and remote sensing imagery as predictors. In most of these studies, one or two machine learning algorithms are chosen, and a variety of evaluation techniques are employed, with most focused on evaluation in one or two unseen years. Evaluation on unseen years is necessary because the model must generalize to weather, farming conditions, and crop distributions that may be different from the years it was trained on. Cai et al. [6] used a multilayer perceptron neural network with three hidden layers; this achieved 95% overall accuracy in differentiating between corn and soybeans by late June in Illinois in unseen years 2014 and 2015. For an end-of-season version of their model, they also provided evaluations on unseen years 2006 through 2015.

Some studies focused on generalizability to unseen regions rather than to unseen years. Blickensdörfer et al. [7] used a random forest for end-of-season crop mapping across Germany, achieving a mean F1 score of 0.79 for corn in 2017 through

Research was sponsored in part by the United States Army Corps of Engineers (USACE) Engineer Research and Development Center (ERDC) Geospatial Research Laboratory (GRL) and was accomplished under Cooperative Agreement Federal Award Identification Number (FAIN) W9132V-22-2-0001. The views and conclusions contained in this document are those of the authors and should not be interpreted as representing the official policies, either expressed or implied, of USACE EDRC GRL or the U.S. Government. The U.S. Government is authorized to reproduce and distribute reprints for Government purposes notwithstanding any copyright notation herein.

Corresponding author: August Posch.

August Posch is with the Institute for Experiential Artificial Intelligence and Sustainability and Data Sciences Lab, Northeastern University. 216 Massachusetts Avenue, Boston, MA 02155 USA (e-mail: posch.au@northeastern.edu).

Jitendra Kumar is with the Environmental Sciences Division, Oak Ridge National Laboratory, 1 Bethel Valley Road, Oak Ridge, TN 37830 USA (email: kumarj@ornl.gov).

Forrest M. Hoffman is with the Computational Science and Engineering Division, Oak Ridge National Laboratory, 1 Bethel Valley Road, Oak Ridge, TN 37830 USA (email: hoffmanfm@ornl.gov).

Auroop R. Ganguly is with the Institute for Experiential Artificial Intelligence and Sustainability and Data Sciences Lab, Northeastern University. 216 Massachusetts Avenue, Boston, MA 02155 USA (e-mail: a.ganguly@northeastern.edu).

Reproduce this work with open-source code: https://github.com/augustposch/crops_13Aug2024

2019 when validating on held-out regions within the previously-seen years from training. Similarly, Sharma et al. [8] used random forest, support vector machine, and XGBoost models to achieve a 0.92 F1 score for corn in Iowa by end of July when training and validating on the same year 2021. Li, Haijun et al. [9] created end-of-season crop type maps for China using Sentinel and PlanetScope remote sensing products and random forest models. Their approach involved an initial field-level step before the pixel level, and they used 10-fold spatial cross validation to report metrics across all geographies, achieving an F1 score of 0.86 for corn end-of-season across China in the year 2019.

The in-season crop mapping studies that provide the most thorough evaluation might use several unseen years for validation and might use multiple algorithms. You et al. [10] used a Gaussian mixture model to achieve an 0.89 F1 score among croplands for mapping corn in Iowa by early August in unseen years 2019 and 2020. Konduri et al. [11] pioneered a cluster-then-label and ecoregion-based approach with the *Mapcurves* [12] algorithm; with robust validation on four unseen years 2015-2018, they achieved a mean F1 score of 0.64 for corn end-of-season across CONUS, with early-June scores 0.10 lower. Johnson and Mueller [13] used a random forest model to create maps from both crop rotation history and remote sensing data from Landsat and Sentinel products. In the unseen year 2020, their model achieved an F1 score of 0.88 for corn at end-of-season across the US Corn Belt, achieving F1 of 0.56 for corn by early June. In another approach, Zaheer et al. [14] created in-season crop type maps using Landsat and Sentinel remote sensing and a convolutional neural network (CNN) U-Net model. They achieved an F1 score of 0.88 for corn in the US Corn Belt by the end of August when validated on the previously-seen years 2017-2019 from training; for unseen years 2020-2021, they also reported overall accuracy of 92% for Iowa by end of August, although crop-specific accuracies were not reported and the same study found corn F1 score was 5 points below overall accuracy across the Corn Belt. Li, Hui et al. [15] created a CONUS-wide in-season crop type map at 30m resolution using a random forest model and Landsat and Sentinel remote sensing data in conjunction with a trusted pixel method that indirectly takes advantage of crop rotation history. They achieved an F1 score of 0.90 by May for Iowa in the unseen year 2022.

In summary, many algorithms have been studied, including tree-based algorithms like random forest and XGBoost, neural networks like MLP and CNN U-Net, and methods involving unsupervised techniques like Gaussian Mixture Model and cluster-then-label *Mapcurves*. It is desirable to compare amongst all the methods, but the above studies used a variety of remote sensing data, resolutions, processing schemes, accuracy metrics, and validation years. To make a fair comparison across algorithms, there is a need to hold those aspects fixed.

In order to understand the relative performance of algorithms, the validation technique must be standardized and account for interannual uncertainty. Plants have varying phenology year to year due to weather, soil and farming decisions, leading crop classification difficulty to vary depending on the year. Therefore, not only do practitioners need all algorithms to be validated in the same year or years, but also practitioners need accuracy metrics over *many* years to ensure that a reasonable range of weather conditions has been covered. Finally, the in-season crop mapping problem demands evaluation in unseen years, because operationally, farmers and policymakers will be using the model in a year it wasn't trained on. Putting these aspects together, it is imperative that all models are evaluated on the same set of many unseen validation years.

The present study focuses on systematic evaluation of a wide range of algorithms with an operationally-relevant validation scheme. This provides value for farmers, emergency managers, and policymakers. Results are presented for corn and almond crop types in Iowa and California respectively, in order to quantify algorithm performance on select important crop types grown within different farming contexts. Each model receives predictor data from crop rotation history and Harmonized Landsat-Sentinel (HLS) remote sensing imagery, in an attempt to give all models the best chance at high accuracy. Finally, an investigation of why certain algorithms succeed in certain conditions is conducted to provide practitioners further trust in the methods.

The main contributions of the present study are 1) thorough intercomparison of a wide range of machine learning-based methods for in-season crop type mapping, and 2) evaluation across five unseen years to report performance in a way that accounts for interannual variability and quantifies uncertainty.

The main research goals are to 1) systematically compare many machine learning algorithms and configurations against each other to understand their strengths and weaknesses; and 2) robustly validate performance to quantify the range of performance to expect when applied to an unseen year.

## II. Methods

### A. Datasets and study area

The Cropland Data Layer (CDL) is a crop type map product at 30m resolution in Albers Conic Equal Area projection (EPSG:5070) covering the entire CONUS [5]. It is an end-of-season product typically available the February after harvest. The product is developed by USDA via machine learning on remote sensing data, using confidential Farm Service Agency on-the-ground surveys as ground truth. For corn in Iowa, both the user accuracy and producer accuracy are roughly 96% for every year 2018-2022; for almonds in California, both the user accuracy and producer accuracy are roughly 90% for every year 2018-2022.

The Harmonized Landsat-Sentinel (HLS) product is developed by NASA to combine observations from two different programs to provide a single dense time series of multispectral surface reflectance imagery [16], [17]. Landsat is a long-running NASA program that, in 2013, launched its Landast-8 satellite, which records surface reflectance at 30m resolution. Sentinel-2 is a program from the European Space Agency, with the Sentinel-2A satellite launched in 2015 and the Sentinel-2B satellite launched in 2017. Both Sentinel satellites collect surface reflectance at 10m, 20m, or 60m resolution, depending on the spectral band. The HLS project runs algorithms that spectrally adjust Sentinel data to match the Landsat OLI instrument and regrid Landsat to fit the Sentinel-2 gridding scheme but at 30m resolution. The combination of Landsat-8, Sentinel-2A, and Sentinel-2B provides a dense time series with observations every 1-2 days, ultimately providing scientists more cloud-free information about what's on the ground. The HLS product is provided in 30m spatial resolution in UTM projection, with global land coverage and fifteen spectral bands. This study uses six of its spectral bands: blue, green, red, near-infrared (NIR), shortwave infrared-1 (SWIR1), and shortwave infrared-2 (SWIR2).

Two study areas were selected: a 100km-by-100km region encompassing Sacramento, California, where almonds are the focus crop, and a 100km-by-100km region encompassing Ames, Iowa, where corn is the focus crop (Figure 1). These were selected due to the CDL's high crop-region accuracies, as well as the unique aspects of these crops and regions. Corn is the most planted crop by acreage in the U.S. and in Iowa, and is a vital part of food supply, animal feed, and ethanol fuel. In the Ames study region, during 2018-2022, corn covers 36.1% to 41.1% of the CDL pixels, figures rivaled only by soybeans at 27.0% to 30.3%. In California, almonds are one of the most common crops, and 80% of the world's almond supply is produced there. However, almonds are one of many common crops in the Sacramento study region in 2018-2022, covering only 1.0% to 3.5% of the CDL pixels, with similar figures for walnuts, grapes, tomatoes, alfalfa, corn, and winter wheat. The small, phenotypically diverse farms in California provide one type of challenge for a machine learning algorithm, while in Iowa the dominance of corn and soybeans, which are phenotypically very similar to each other, provides another type of challenge for the machine learning algorithm.

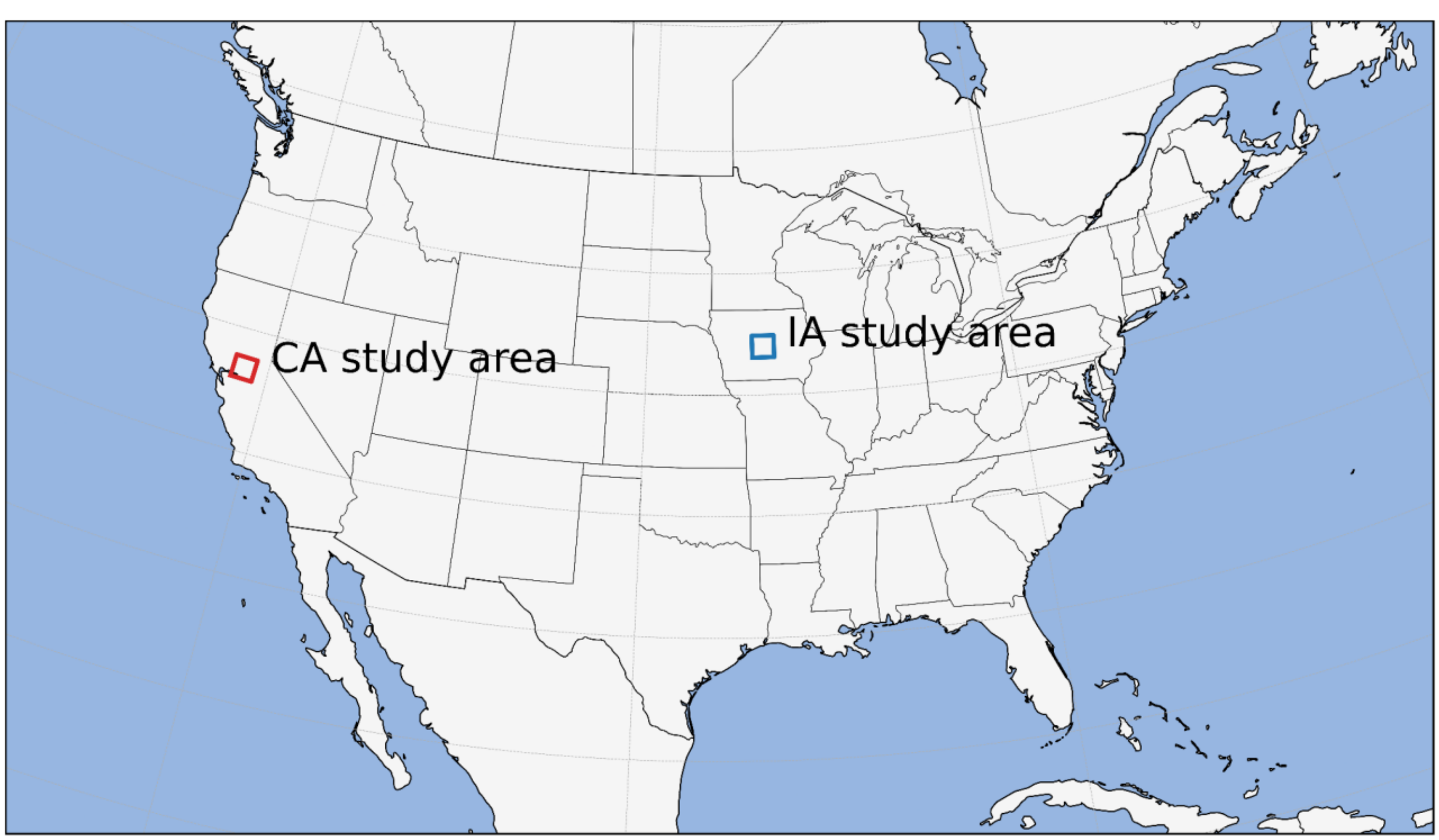

Figure 1. Location of the two study areas within the United States. The California (CA) study area is marked by the red square. The Iowa (IA) study area is marked by the blue square. Both study areas are 100 km by 100 km.

*B. Data processing steps*

National CDL data for years 2014-2022 and HLS data for the two regions of interest for years 2018-2022 were obtained and processed. To process HLS and CDL to a common grid, the CDL was reprojected into UTM10N for the California study area and into UTM15N for the Iowa study area, using nearest-neighbor resampling to determine the crop label.

HLS data were filtered to select only the unobstructed observations. For clouds and cloud shadows, the HLS product's built-in QA band was used. For snow, a filter was written to flag pixels with blue reflectance exceeding 0.2. All such cloud, cloud shadow, and snow observations were discarded. The rest of the reflectance data, believed to be unobstructed observations of the surface, were used in the temporal compositing step.

Temporal compositing was performed on the unobstructed reflectance data for each pixel location, year, and band. The year was divided into periods, and the median reflectance value within each period was taken. Two different compositing schemes were used to partition the year, because shorter (5-day) compositing periods may track phenological progress more finely, while longer (14-day) compositing periods may produce fewer gaps and reduce the effect of outliers. Both schemes have extra-long periods early (days of year 1-90) and late (days of year 301-366) in the calendar year to avoid snow and cloud issues. The 5-day scheme has 44 periods in a year, composed of days-of-year 1-90, 91-95, 96-100, 101-105, … , 296-300, and 301-366. The 14-day scheme has 17 periods, composed of days-of-year 1-90, 91-104, 105-118, 119-132, … , 287-300, and 301-366.

Henceforth reflectance values are referred to as "features" for the machine learning analysis. Using the 14-day compositing scheme, at each pixel one year has 17 periods and 6 bands, and thus 102 reflectance features. Similarly, for the 5-day scheme, one year has 44 periods and 6 bands, providing 274 reflectance features at each pixel.

Data gaps due to obstruction from clouds, cloud shadows, and snow were analyzed before performing interpolation to fill the gaps. For Iowa region 14-day datasets, in every year at least two-thirds of the pixels had ample data, with no more than 18 missing features out of 102 features. For California region 14-day datasets, in every year at least 99% of the pixels had ample data, with no more than 18 missing features out of 102 features. For California region 5-day datasets and for Iowa region 5-day datasets, in every year all the pixels had less than half their 274 features missing. Missing values had a spiky distribution in time, but the spikes were spread uniformly across the year, suggesting that these gaps could easily be filled by interpolation.

Because the data gaps were small enough, cubic spline interpolation was used to fill the missing values. A cubic spline was fitted for each pixel, each year, and each band across the growing season (days-of-year 91-300). Any missing-data periods were filled with interpolated values from the cubic spline.

Finally, datasets for machine learning were created. With 2 study areas, 5 years, and 2 compositing schemes, there are 20 datasets, each structured as follows. Each row (or "data point") represents one pixel in one year, so there are 13 million data points. Each column (or "feature") represents reflectance for one period for one band, so the 14-day scheme has 102 reflectance features and the 5-day scheme has 274 reflectance features. The early-season prediction problem is defined using a subset of the features. For example, to predict a crop type knowing only its pre-early-June reflectance signature in the 14-day scheme, the algorithm sees only information from the one 90-day winter period and the first five 14-day periods, thus 6 periods and 6 bands providing 36 reflectance features.

Crop rotation history features were also added to the input datasets, using previous years of CDL. For each year's reflectance dataset, four years of crop rotation history features were added – 1 feature for each of the prior 4 years, a Boolean of whether that pixel is the crop of interest or not. Hence, in the pre-early-June example with the 14-day scheme, the algorithm sees 40 features in total, comprising the 36 reflectance features and the 4 crop rotation history features.

Finally, the label $y$ was produced from the CDL in the year matching the reflectance data. The label is a Boolean of whether that pixel is the crop of interest or not.

For machine learning algorithms that required scaled data, the features were standard-scaled so that each feature had mean 0 and variance 1.

### *C. Algorithm intercomparison and hyperparameter search*

Ten machine learning algorithms were explored. These ten specifically were selected because they represent a wide range of decision boundary shapes and complexities, and due to their scikit-learn implementations, they are readily available for practitioners. For each algorithm, a grid search of hyperparameters was performed. The choice of which hyperparameters to adjust and the exact hyperparameter values to search over were determined through a mix of trial and error and machine learning literature, while also remaining somewhat efficient with computational resources.

Hyperparameters determine the type of decision boundary the learning algorithm is aiming for. For example, in logistic regression, important hyperparameters to specify include *solver*, *C*, *class_weight*, *penalty*, and *elasticnet*. The 'saga' solver was selected for its efficiency on these datasets and its allowance for elastic net regularization penalty. Regularization is a way to force a simpler model, i.e., forcing model parameters closer to 0, and the strength of regularization is controlled by the *C* hyperparameter, where high *C* means a less-regularized model and low *C* means a more-regularized model. The type of regularization is controlled by the *penalty* hyperparameter: an L1 penalty means "closer to 0" is measured with Manhattan distance, while an L2 penalty means "closer to 0" is measured with Euclidean distance, and an elastic net penalty allows for some compromise between L1 and L2 penalty. When using an elastic net penalty there is an additional hyperparameter known as *l1_ratio* controlling whether the compromise leans towards L1 or leans towards L2. Other algorithms use some similar hyperparameter concepts as the logistic regression, but each has its own unique set to search over. Below we introduce the ten algorithms and their hyperparameters.

1. Random Forest (RF) is an algorithm which is an ensemble of many decision trees [18]. Each individual decision tree is fit to a subset of the training data using a series of best possible splits, and predictions are made using a majority vote from the entire forest of trees. This study used scikit-learn's [19] RandomForestClassifier, searching over hyperparameters n_estimators in [200, 500], max_features in [0.05, 0.1, 0.2, 0.4, 1.0], min_samples_split in [2, 4], bootstrap in [False, True], and class_weight in [None, 'balanced'], with all other parameters kept at defaults, for a total of 80 hyperparameter combinations.
2. Extra Trees (ET), or extremely-randomized trees, is an algorithm which is an ensemble of many decision trees [20]. Each individual decision tree is fit to a subset of the training data using a series of splits based on random thresholds, and predictions are made using a majority vote from the entire forest of trees. This study used scikit-learn's ExtraTreesClassifier, searching over hyperparameters *n_estimators* in [200, 500], *max_features* in [0.05, 0.1, 0.2, 0.4, 1.0], *min_samples_split* in [2, 4], *bootstrap* in [False, True], and *class_weight* in [None, 'balanced'], with all other parameters kept at defaults, for a total of 80 hyperparameter combinations.

3. Logistic Regression (LR) is an algorithm that learns a coefficient for each feature determining how much that feature contributes to a 1 or 0 prediction (e.g., Almonds or Not Almonds) [21]. Without regularization, the maximum likelihood coefficients are learned, but with regularization the algorithm is encouraged to learn coefficients closer to 0. This study used scikit-learn's LogisticRegression with *solver* 'saga' [22], searching over hyperparameters *C* in [1, $10^{-2}$, $10^{-4}$, $10^{-6}$, $10^{-8}$], *class_weight* in [None, 'balanced'], *penalty* in ['elasticnet', None], and *elasticnet* in [0, 0.1, 0.5, 0.9, 1], for a total of 52 hyperparameter combinations.
4. Linear Support Vector Machine (LS) is an algorithm that learns a linear decision boundary that maximizes the margin between the boundary and points of either class [23]. When the two classes cannot be perfectly separated, allowances are made for data points that are on the wrong side of the boundary, and the flavor and permissiveness of this regularization is specified by hyperparameters. In this study, scikit-learn's LinearSVM was used, searching over hyperparameters *penalty* in ['l1', 'l2'], *loss* in ['hinge', 'squared_hinge'], C in [1, $10^{-2}$, $10^{-4}$, $10^{-6}$, $10^{-8}$], and *class_weight* in [None, 'balanced'] for a total of 30 hyperparameter combinations.
5. Nonlinear Support Vector Machine (SV) learns the boundary that maximizes the margin between points of either class and the boundary [23]. Again, allowances are made for data points on the wrong side of the boundary and the form of this regularization is determined by hyperparameters. Kernels transform the input space to allow the classes to be separable (e.g., allow us to separate almonds from non-almonds) in a higher-dimensional space. The kernel can be a radial basis function, which transforms the input space according to Gaussians centered on each point; polynomial, which transforms the input space according to a polynomial function fitted to each class; or sigmoid, which transforms the input space in ways similar to those used in neural networks. The gamma hyperparameter adjusts the reach of the kernels; for example, higher gamma in radial basis function makes the Gaussians have low variance, so the classification is more like a 1-nearest-neighbor classifier, whereas lower gamma in radial basis function makes the Gaussians have high variance, so the classification is more influenced by farther-away training data points. This study used scikit-learn's SVM with *degree* always set to 3, searching over hyperparameters *kernel* in ['rbf','poly','sigmoid'], *gamma* in [100, 10, 1, $10^{-1}$, $10^{-2}$], and *C* in [1, $10^{-2}$, $10^{-4}$, $10^{-6}$, $10^{-8}$], for a total of 150 hyperparameter combinations.
6. Gaussian Naïve Bayes (NB) is an algorithm that represents each training data point by a multidimensional Gaussian probability density function in the feature space, resulting in a probability density function for each class, and then to classify a new data point the Bayes Theorem is used and the density of each class is compared [24]. The Naïve part is that each feature dimension is modeled independently, so the Gaussians have no covariance, which makes the Bayesian calculations tractable. The *var_smoothing* hyperparameter adds slightly more variance to the Gaussians in all dimensions to ensure better extension to unseen data outside the training distribution. This study used scikit-learn's GaussianNB, searching over *var_smoothing* in [$10^{-6}$, $10^{-7}$, $10^{-8}$, $10^{-9}$, $10^{-10}$, $10^{-11}$, 0], for a total of 7 hyperparameter combinations.
7. K Nearest Neighbors (abbreviated KN in this study) is an algorithm that, for each new data point, looks for the most similar (nearest in feature space) data points in the training set and uses the majority class of those as the prediction of the class of the new data point [25]. The distance metric that defines what nearest means, the number of neighbors allowed, and whether to weight points by their distance are all hyperparameters that can be tuned. In this study, scikit-learn's KNeighborsClassifier was used, searching over 19 values of *n_neighbors* exponentially spread between 1 and 1000, *weights* in ['uniform', 'distance'], and *metric* in ['manhattan', 'euclidean', 'chebyshev', 'canberra'], for a total of 152 hyperparameter combinations.
8. Neural Network Multilayer Perceptron (NP) is an algorithm that learns the best weights and biases for every connection of layers of hidden nodes between input features and output class prediction [26], [27]. Each individual connection has a simple activation function, but the dense connections together allow the entire network to approximate a wide variety of function shapes. There are many types of neural networks, but for this study, networks with 1 to 3 hidden layers were tried, where all nodes of each layer are connected to all nodes of each following layer, and this form of neural network is known as a Multilayer Perceptron (MLP). In this study, scikit-learn's MLPClassifier was used, searching over hyperparameters *hidden_layer_sizes* in [(100,), (100,100,100)], *activation* in ['relu', 'logistic', 'tanh'], and *alpha* in [1, $10^{-2}$, $10^{-4}$, $10^{-6}$, $10^{-8}$, $10^{-10}$], for a total of 36 hyperparameter combinations.
9. Principal Components Analysis (PCA) is an

unsupervised machine learning technique that represents a high-dimensional dataset in a lower-dimensional way while preserving as much of its variance as possible [28], [29]. In this study it was used as an initial dimensionality reduction step in Principal Components K Nearest Neighbors (abbreviated PK in this study) and in Principal Components Logistic Regression (abbreviated PL in this study). It was used in these ways because K Nearest Neighbors and Logistic Regression are known to sometimes perform better with fewer input dimensions. In each case, PCA was performed using scikit-learn's PCA, with the *n_components* hyperparameter always set to 0.9, telling the algorithm to learn a new representation of the features such that 90% of variance is preserved, but using the fewest possible number of feature dimensions. For PK, the new representation was sent to the KNeighborsClassifier, and the same hyperparameters were searched over as for KN, for a total of 152 hyperparameter combinations in PK. For PL, the new representation was sent to the LogisticRegression classifier, and the same hyperparameters were searched over as for LR, for a total of 52 hyperparameter combinations in PL.

10. As mentioned above, Principal Components Analysis was used similarly in Principal Components K Nearest Neighbors (PK) and in Principal Components Logistic Regression (PL), thus bringing the total to ten algorithms.

In addition to all the hyperparameter combinations, there were 12 scenarios, representing different compositing schemes (5-day or 14-day), different in-season cutoffs (early June, mid-August, end of year), and different crops and regions (Almonds in California or Corn in Iowa). Thus, with 791 hyperparameter configurations running on 12 scenario configurations, there were 9492 models that were sent through the cross-validation process. Model configurations are named according to the relevant attributes (Table I and Table II), for example, "RF079-5day" refers to a Random Forest model with the 79th hyperparameter combination and the 5-day HLS compositing scheme.

Table I
ALGORITHMS TRAINED IN THE STUDY.

| **Algorithm** | **Code** |
|---|---|
| Random Forest | RF |
| Extra Trees | ET |
| Logistic Regression | LR |
| Linear Support Vector Machine | LS |
| Nonlinear Support Vector Machine | SV |
| Gaussian Naïve Bayes | NB |
| K Nearest Neighbors | KN |
| Neural Network Multilayer Perceptron | NP |
| Principal Components K Nearest Neighbors | PK |
| Principal Components Logistic Regression | PL |

Table II
RESULTS FOR EACH ALGORITHM. CONFIGURATIONS ARE NAMED ACCORDING TO THEIR ALGORITHM AND THEIR HYPERPARAMETER COMBINATION, SO "RF079-5DAY" IS THE RANDOM FOREST WITH THE 79TH HYPERPARAMETER COMBINATION AND THE 5-DAY COMPOSITING SCHEME. SEPARATE COLUMNS OF RESULTS ARE PROVIDED FOR CALIFORNIA ALMONDS PREDICTION BY JUNE AND FOR IOWA CORN PREDICTION BY JUNE.

| | | California almonds June | | Iowa corn June | |
|---|---|---|---|---|---|
| **Algorithm** | **Code** | **Best configuration** | **Mean F1 score** | **Best configuration** | **Mean F1 score** |
| Random Forest | RF | RF079-5day | 0.723 | RF067-14day | 0.566 |
| Extra Trees | ET | ET014-14day | 0.725 | ET046-14day | 0.577 |
| Logistic Regression | LR | LR036-14day | **0.739** | LR043-14day | 0.563 |
| Linear | LS | LS001-14day | **0.739** | LS021-14day | **0.593** |
| Support Vector Machine | | | | | |
| Nonlinear Support Vector Machine | SV | SV009-14day | 0.732 | SV088-14day | 0.587 |
| Gaussian Naïve Bayes | NB | NB003-14day | 0.621 | NB007-14day | 0.465 |
| K Nearest Neighbors | KN | KN079-14day | 0.735 | KN091-14day | 0.579 |
| Neural Network Multilayer Perceptron | NP | NP005-5day | 0.710 | NP003-14day | 0.579 |
| Principal Components K Nearest Neighbors | PK | PK041-14day | 0.663 | PK143-14day | 0.569 |
| Principal Components Logistic Regression | PL | PL003-5day | 0.657 | PL033-14day | 0.567 |

*D. Evaluation*

Mathematically, this study performs Empirical Risk Minimization to find the best function from features to crop label.

In machine learning, the goal is to approximate a function $f: X \to Y$ using a training set $\{x_i, y_i\}_{i=1}^{N}$ . This study poses a classification problem – e.g., a binary classification of Almonds or Not Almonds – so $Y = \{0,1\}$ where 1 indicates Almonds. Likewise, $X$ is the set of pixels in a given year. Given a pixel $x_i \in X$, the function $f$ assigns it the correct label $y_i \in Y$ – in other words, $f$ is the hypothetical perfect function that is always correct.

The machine learning algorithm finds a candidate function $\hat{f}$ that generalizes well from finite training data $S$ to the infinite space $X \times Y$ of all potential data. To measure how closely the candidate function predicts the true label, some accuracy metric is defined $\text{Acc}(y, f(x))$ . To capture "generalizes well", the ideal goal is to find, out of *all possible* functions $\mathcal{F}$, the function $f^*$ that maximizes expected accuracy on *all potential data*:

$$f^* = \underset{f \in \mathcal{F}}{\text{argmax}}\, \mathbb{E}_{(x,y) \in X \times Y} \left[\text{Acc}\big(y, f(x)\big)\right] \quad (1)$$

In practice, only a subset of functions $F \subset \mathcal{F}$ can be considered (in this study, $F$ is the set of hundreds of kinds of functions described in Methods section C), and only a finite set of data $S$ is available. Hence, the practical goal is to find, out of the *chosen* kinds of functions, which function maximizes *empirical* expected accuracy:

$$f^* = \underset{f \in F}{\text{argmax}}\, \mathbb{E}_{(x_i,y_i) \in S} \left[\text{Acc}\big(y_i, f(x_i)\big)\right] \quad (2)$$

This study aims to find the above $f^*$ for the crop classification problem, by quantifying model accuracy when trained on existing data but predicting on new data.

Uncertainty quantification is also critical for confidence in future predictions. The quantification of accuracy and uncertainty provides confidence in the ability to create the next in-season crop map.

A 5-fold cross validation scheme - where each year served as a validation year for a different fold – provided estimates of model accuracy on unseen data, with uncertainty quantification. Thus in one fold train on 2018-2021, then validate on 2022; in another fold train on 2018-2020 and 2022, then validate on 2021; and so on for five folds. For each fold, precision, recall, and F1 score were recorded, as well as the arithmetic mean and standard error of each across the five folds. The standard errors quantify the uncertainty due to different years of data, which in turn come from different weather and farming decisions in different years. The accuracy metrics are based on the confusion matrix (Table III).

Table III
CONFUSION MATRIX

| | Reference | |
|---|---|---|
| | Yes | No |

| | | | |
|---|---|---|---|
| Prediction | Yes | $a$ | $b$ |
| | No | $c$ | $d$ |

Precision is also known as user accuracy (UA) and success ratio (SR). In a corn-mapping problem, precision measures how many of the model-predicted corn pixels were actually corn in the reference dataset. A model with low precision misleads stakeholders into thinking there is corn where there is no corn.

$$\text{Precision} = \frac{a}{a+b} \tag{3}$$

Recall is also known as producer accuracy (PA) and probability of detection (POD). In a corn-mapping problem, recall measures how many of the reference corn pixels were predicted corn by the model. A model with low recall misleads stakeholders into thinking there is no corn where there is corn.

$$\text{Recall} = \frac{a}{a+c} \tag{4}$$

The F1 score is the harmonic mean of precision and recall, and it places emphasis on both precision and recall being high. Both low precision and low recall are important to avoid, so in this study F1 score was chosen as the primary metric for how well the model predicts where the corn is.

$$\text{F1 score} = \frac{2}{\frac{1}{\text{Precision}} + \frac{1}{\text{Recall}}} \tag{5}$$

Due to the computational requirements of performing cross-validation for 9492 models (1592 models per crop type per cutoff date), two passes of model runs were performed with different amounts of training data. A first pass of model runs was performed using 0.1% uniform random samples of training data in each fold and 0.1% uniform random samples of validation data in each fold. For each of the 9492 model runs, the same 0.1% sample was used, so it was a fair comparison of models. Based on the first pass, the best model configurations were identified and for these best configurations a second pass of model runs was performed using 100% of training data in each fold.

After running all these models through the comparison scheme, all their metrics were analyzed for strengths and weaknesses of each configuration, but the ultimate comparison was made based on mean F1 score across the years.

### *E. Connecting hyperparameters to interannual patterns for California almonds by early June*

The large set of algorithm outputs was further investigated to discern interannual phenological variation and its effect on classification accuracy, with an eye towards which hyperparameter settings are suitable for which kinds of years. These analyses were performed only for the California almonds study area, for a few reasons: almonds are under-studied compared to corn; the best almond models showed greater interannual variation than the best corn models; and the top almond models came from four of the ten algorithms, a wider representation than the top corn models, which came from only two of the ten algorithms.

To begin this interannual analysis of California almond models, their performance scores were plotted for pairs of validation years, and Pearson correlations between scores for pairs of years were calculated. This was done both for all 1592 California almonds early June models, and again separately for those that achieved a mean F1 score within 0.01 of the best. Prevalence of almonds in each year was calculated, and rate of crop switching was calculated. Then the feature space was characterized by calculating feature correlations and separating reflectance patterns from crop rotation history patterns.

As a further case study, K Nearest Neighbors hyperparameter combinations were investigated for mean performance across years and for performance in particular years. K Nearest Neighbors was showcased because its hyperparameters are easy to visualize, and it produced one of the top models for almond mapping. The deeper analysis of KN algorithm settings serves as demonstration of best practices in crop mapping algorithm selection.

Together, the above evaluations generated insights about why certain years were more challenging for almond classification.

### *F. Computing*

Analyses were developed using Python, with NumPy and Pandas for data array manipulation, PyGDAL and Rasterio for geospatial and raster manipulation, SciPy for data processing, and Scikit-learn for machine learning.

Running 9492 model configurations through a cross-validation scheme required ample computation, so algorithms were run on anywhere from 20 to 120 CPUs in parallel. Massachusetts Green High Performance Computing Center (MGHPCC) resources were used, along with the Dask library for parallel computing in Python. The Dask cluster was configured optimally for use on MGHPCC. The 0.1% sample datasets were pre-processed in a separate step before the model runs in order to save memory among the threads running the models.

To run the model training with 100% samples in the second pass, a bagged classifier approach was used to overcome computational limitations. Bagging computes many (say, 100 or 1000) different instances of the same-hyperparameter algorithm, each computed on different ("bagged") random samples of the input data, and then these instances all vote on the classification [30]. The bagging technique allows the whole dataset's information to be used while keeping each individual instance small.

## III. Results

### *A. Performance rankings by algorithm*

All hyperparameterizations of each algorithm were evaluated based on mean F1 score across validation years. The best mean

F1 score for each algorithm (further subdivided by crop type, region, and in-season date) are the results reported in Table II. For example, the Random Forest performance in Table II comes from the best hyperparameterization of Random Forest based on mean F1 score. Note also that visual representations of mean performance metrics are available in Figures 2 and 3, and visuals with mean and standard error of F1 score are provided in supplementary figures S1 and S2.

For almonds in California by early June (Figure 2), the RF, ET, LR, LS, SV, and NP algorithms all achieved mean F1 scores above 0.71. LR and LS models achieved the very best mean F1 scores, at 0.74, and these are discussed further below. NB, PK, and PL models were unable to perform as well, only achieving mean F1 scores of 0.62 to 0.66. Since NB was the simplest algorithm considered and PK and PL both simplify the feature space via PCA, the results suggest that the almond-mapping problem requires algorithms that allow for more complexity.

For corn in Iowa by early June (Figure 3), nine of the ten algorithms were able to achieve a mean F1 score above 0.56. The exception was the NB algorithm, at 0.47. The very best mean F1 scores came from SV and LS algorithms, at 0.59, discussed further below.

*B. Best algorithms for almonds in California by early June*

The best algorithm for almonds in California by early June was either Logistic Regression or Linear SVM, tied with a mean F1 score of 0.74. Figure 2 shows that these two algorithms performed the best. The mean precision across years was 0.73, meaning 73% of the early-June-mapped almond pixels are actually almonds in the CDL. The mean recall across years was 0.83, meaning 83% of the almond pixels in the CDL appear as almonds in the early-June map. There were 15 configurations that were tied for the best mean F1 score of 0.7389, and they all had identical metrics, so here we report configurations of one LR and one LS out of the 15. Specific configurations were "LS021-14day", and "LR036-14day". "LS001-14day" has class weights none, type of loss squared hinge, type of regularization l1, and strength of regularization C=1, and was run on the 14-day composited dataset. "LR036" has class weights balanced, type of regularization elastic net with l1 ratio 0.9, strength of regularization C=0.0001, and was run on the 14-day composited dataset. For performance of these models, see Table IV.

Out of the 1582 California almonds early-June models, 38 of them achieved a mean F1 score within 0.01 of the best (i.e. f1_mean>0.7289). In addition to the LS and LR models mentioned above, the top 38 included K Nearest Neighbors models and a nonlinear SVM model. The best KN model was KN079-14day, with mean F1 score of 0.7346. Its hyperparameter settings were distance metric Chebyshev, K number of neighbors 41, weights by distance, and was run on the 14-day composited dataset. The best SV model was SV009-14day, with mean F1 score of 0.7325. Its hyperparameter settings were class weights none, kernel type RBF, kernel width gamma=0.01, strength of regularization C=1, and was run on the 14-day composited dataset. Aside from the four algorithms LS, LR, KN, and SV, no other algorithms reached within 0.01 of the best mean F1 score.

The best California almonds early-June models all had very similar F1 standard error. LS001-14day and LR036-14day had F1 standard error of 0.1506, which was the largest standard error of the aforementioned top 38 models. The smallest F1 standard error of the top 38 was 0.1310. Hence, none of the best models had a standard error meaningfully better than the others. Furthermore, the distribution of performance across validation years has the same shape for all 38: LS001-14day's F1 scores for years 2018-2022 were 0.69, 0.88, 0.47, 0.83, 0.83 respectively; all of the other 38 were no more than 0.03 better in every year. Therefore, in any given year, the performance gain from switching among the top 38 models is far less than the typical year-to-year variation. However, small gains may be possible in each year, as investigated in section E below.

Table IV
PERFORMANCE OF BEST MODELS FOR ALMONDS IN CALIFORNIA – "LS001-14DAY" AND "LR036-14DAY" (LINEAR SVM WITH HYPERPARAMETERIZATION 001 AND 14-DAY COMPOSITING; LOGISTIC REGRESSION WITH HYPERPARAMETERIZATION 036 AND 14-DAY COMPOSITING. ONLY INFORMATION THROUGH EARLY JUNE WAS USED.)

| | Precision (User Accuracy) | Recall (Producer Accuracy) | F1 |
|---|---|---|---|
| Validation year 2018 | 0.6000 | 0.8143 | 0.6909 |
| Validation year 2019 | 0.8929 | 0.8741 | 0.8834 |
| Validation year 2020 | 1.0000 | 0.3036 | 0.4658 |
| Validation year 2021 | 0.8153 | 0.8440 | 0.8294 |
| Validation year 2022 | 0.8352 | 0.8155 | 0.8252 |
| Mean | 0.8287 | 0.7303 | 0.7389 |
| Standard Error | 0.1311 | 0.2145 | 0.1506 |

*C. Best algorithms for corn in Iowa by early June*

The best algorithm for corn in Iowa by early June was linear or nonlinear SVM, tied with a mean F1 score of 0.59. Figure 3 shows that these two algorithms performed the best. The mean precision across years was 0.60, meaning 60% of the early-June-mapped corn pixels are actually corn in the CDL. The mean recall across years was 0.62, meaning 62% of the corn pixels in the CDL appear as corn in the early-June map. There were five configurations that were tied for the best mean F1 score of 0.5934, and they all had identical metrics; because these five were all LS models, here we choose one of them to explain in detail. "LS021-14day" has type of loss hinge, type of regularization l2, and strength of regularization C=1, and was run on the 14-day composited dataset. See Table V for performance details.

Out of the 1582 Iowa corn early-June models, 11 of them achieved a mean F1 score within 0.01 of the best (i.e. f1_mean>0.5834). All of the top 11 were either LS or SV models. "SV088-14day" has hyperparameter settings class weights balanced, kernel type polynomial with degree 3, kernel width gamma=0.01, strength of regularization $C=10^{-6}$, and was run on the 14-day composited dataset. See Table VI for detailed performance. SV088-14day is mentioned here because it was the lone SV model in the top 11 group, and it had a much smaller F1 standard error than the ten LS models, indicating more consistent performance across years.

The best Iowa corn early-June models all had very similar F1 standard error, except for SV088-14day, whose accuracy was more consistent across years. LS021-14day had F1 standard error of 0.0484, and of the top 11 models, the LS models all had very similar standard errors, ranging from 0.0462 to 0.0541. However, the SV model had F1 standard error of 0.0183, meaning this model's accuracy was more consistent across validation years. Looking closer at SV088-14day's performance across validation years, its worst year was 2018 with F1 score of 0.5676 and its best year was 2021 with F1 score of 0.6136. Meanwhile, LS021-14day had the same worst and best years but exaggerated differences, with F1 scores of 0.5131 in 2018 and 0.6451 in 2021. Thus, among the top 11 mean-F1 models, SV088-14day performs much more consistently across the years considered in this study.

Table V
PERFORMANCE OF "LS021-14DAY" – BEST MODEL FOR CORN IN IOWA
(LINEAR SVM WITH HYPERPARAMETERIZATION 021 AND 14-DAY COMPOSITING. ONLY INFORMATION THROUGH EARLY JUNE WAS USED.)

| | Precision (User Accuracy) | Recall (Producer Accuracy) | F1 |
|---|---|---|---|
| Validation year 2018 | 0.7264 | 0.3967 | 0.5131 |
| Validation year 2019 | 0.5867 | 0.7035 | 0.6398 |
| Validation year 2020 | 0.5177 | 0.6393 | 0.5721 |
| Validation year 2021 | 0.5984 | 0.6997 | 0.6451 |
| Validation year 2022 | 0.5463 | 0.6577 | 0.5969 |
| Mean | 0.5951 | 0.6194 | 0.5934 |
| Standard Error | 0.0717 | 0.1140 | 0.0484 |

Table VI
PERFORMANCE OF "SV088-14DAY" – BEST NONLINEAR MODEL FOR CORN IN IOWA
(NONLINEAR SVM WITH HYPERPARAMETERIZATION 088 AND 14-DAY COMPOSITING. ONLY INFORMATION THROUGH EARLY JUNE WAS USED.)

| | Precision (User Accuracy) | Recall (Producer Accuracy) | F1 |
|---|---|---|---|
| Validation year 2018 | 0.8217 | 0.4336 | 0.5676 |
| Validation year 2019 | 0.7309 | 0.4867 | 0.5843 |
| Validation year 2020 | 0.6317 | 0.5765 | 0.6029 |
| Validation year 2021 | 0.6238 | 0.6038 | 0.6136 |
| Validation year 2022 | 0.5318 | 0.6115 | 0.5689 |
| Mean | 0.6680 | 0.5424 | 0.5875 |
| Standard Error | 0.0994 | 0.0702 | 0.0183 |

### *D. Early-season versus end-of-season performance*

Early-season inputs offered performance as good as end-of-season inputs. That is, when the model was restricted to information through early June, the best of these early-June models achieved a mean F1 score within 0.01 of the best end-of-year models, when the model was provided the full year's worth of information as inputs. This finding holds true both for almonds in California and for corn in Iowa, shown in Figure 4, which has best models by crop, region, and in-season cutoff date. Because of this result, the focus of the other parts of this paper is on early-June models.

### *E. Size of training dataset*

Training on 0.1% of available data was just as good as training on 100% of data. Nine configurations from the first pass were used here: because LR and LS were tied for best in early June performance, one best version of each was chosen for each crop-region; in addition to those four, another five SV configurations were analyzed because their F1 means were within 0.25 of best for one or both crop-regions, and SV complexity suggests that these, if any, might improve in the 100% pass. Thus, these nine hyperparameter configurations from the first pass were now run in the second pass and trained on 100% of available data using the Bagging approach. With 9 hyperparameter configurations, 2 crop-regions, and 2 compositing schemes, that means 36 models. All of them scored within 0.03 of the mean F1 of the 0.1% versions, and the only improving models were not the very best, so no new-best models were found out of this second pass. Figure 5 shows this result for the 18 models trained on the 14-day composited dataset. The same result holds true for the other 18 models trained on the 5-day composited dataset, not shown.

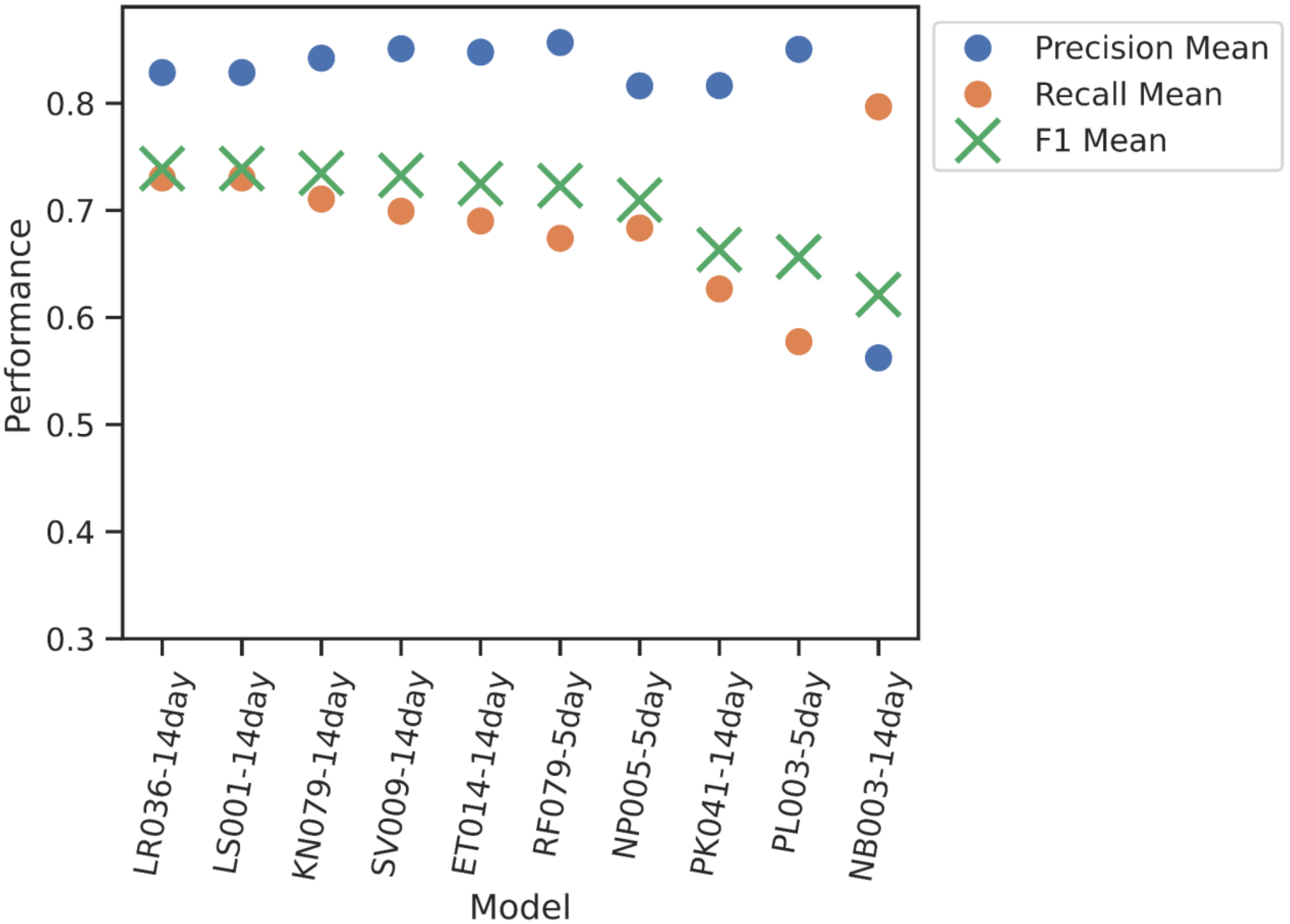

Figure 2. Performance of the best configuration of each algorithm for almonds in California by early June. The best mean F1 scores came from Logistic Regression (LR) and Linear SVM (LS), at 0.74.

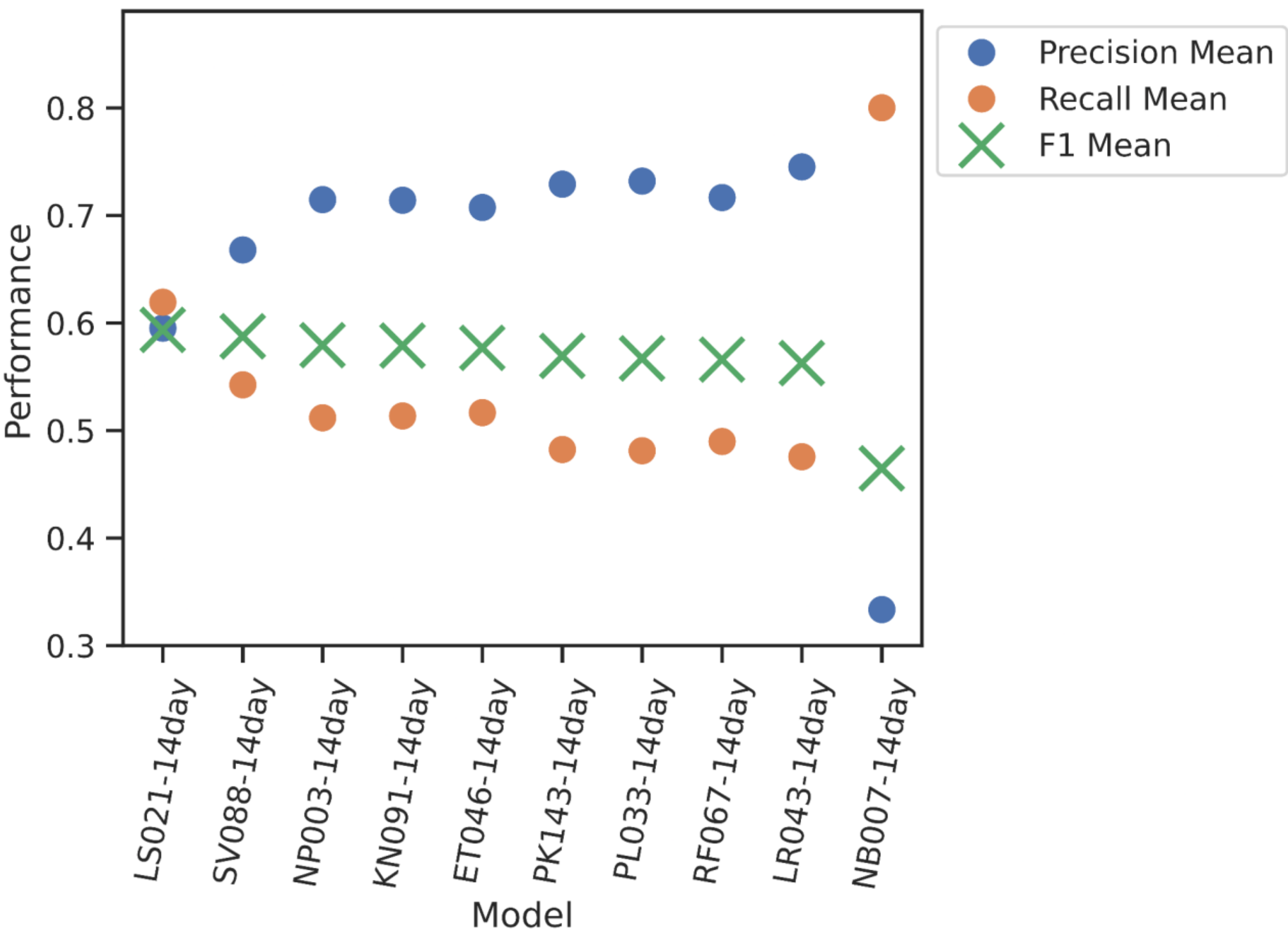

Figure 3. Performance of the best configuration of each algorithm for corn in Iowa by early June. The best mean F1 scores came from Linear SVM (LS) and Nonlinear SVM (SV), at 59%.

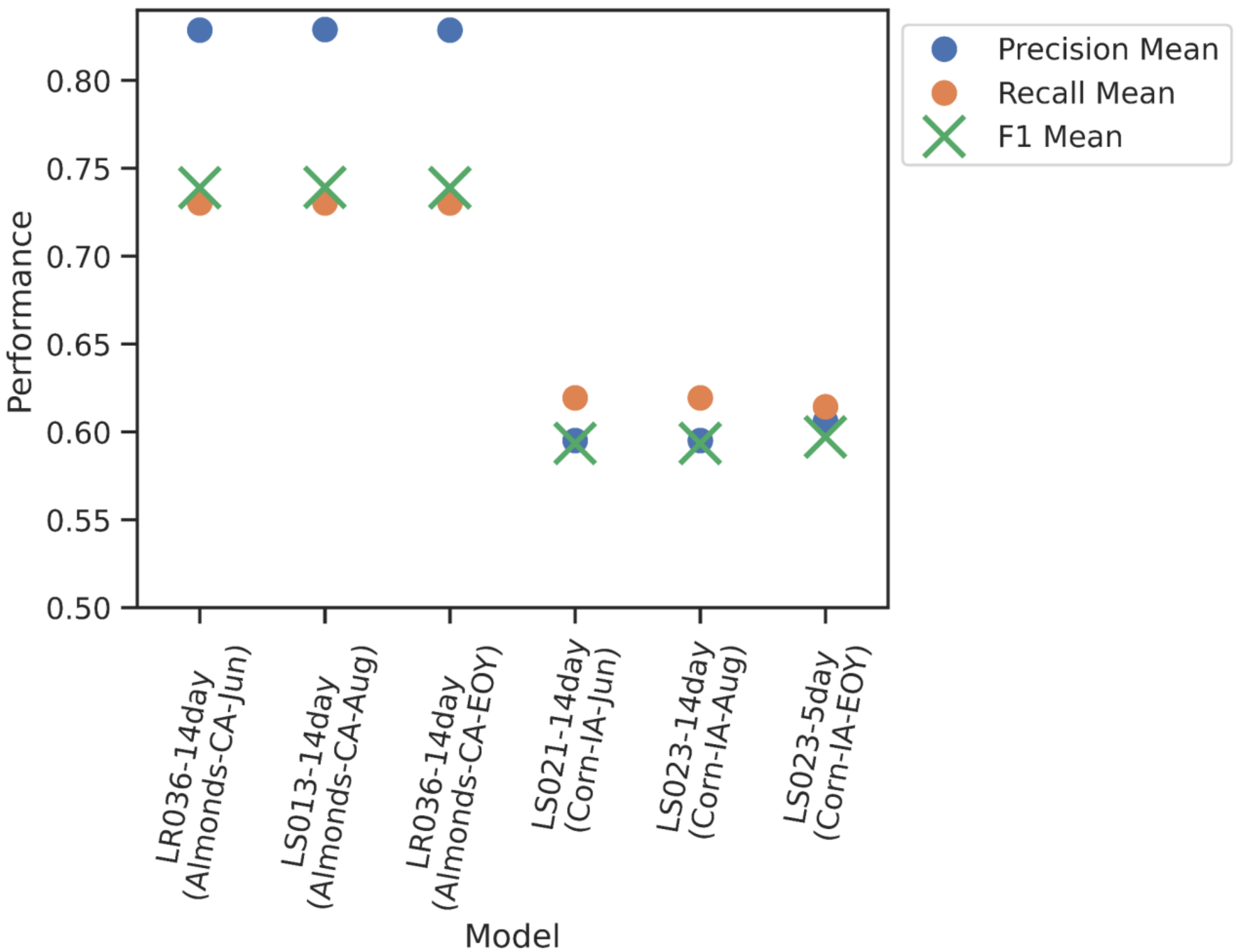

Figure 4. Performance of the best model for each in season cutoff date, such as June, August, or End Of Year. Early-season models achieve performance on par with end-of-season models.

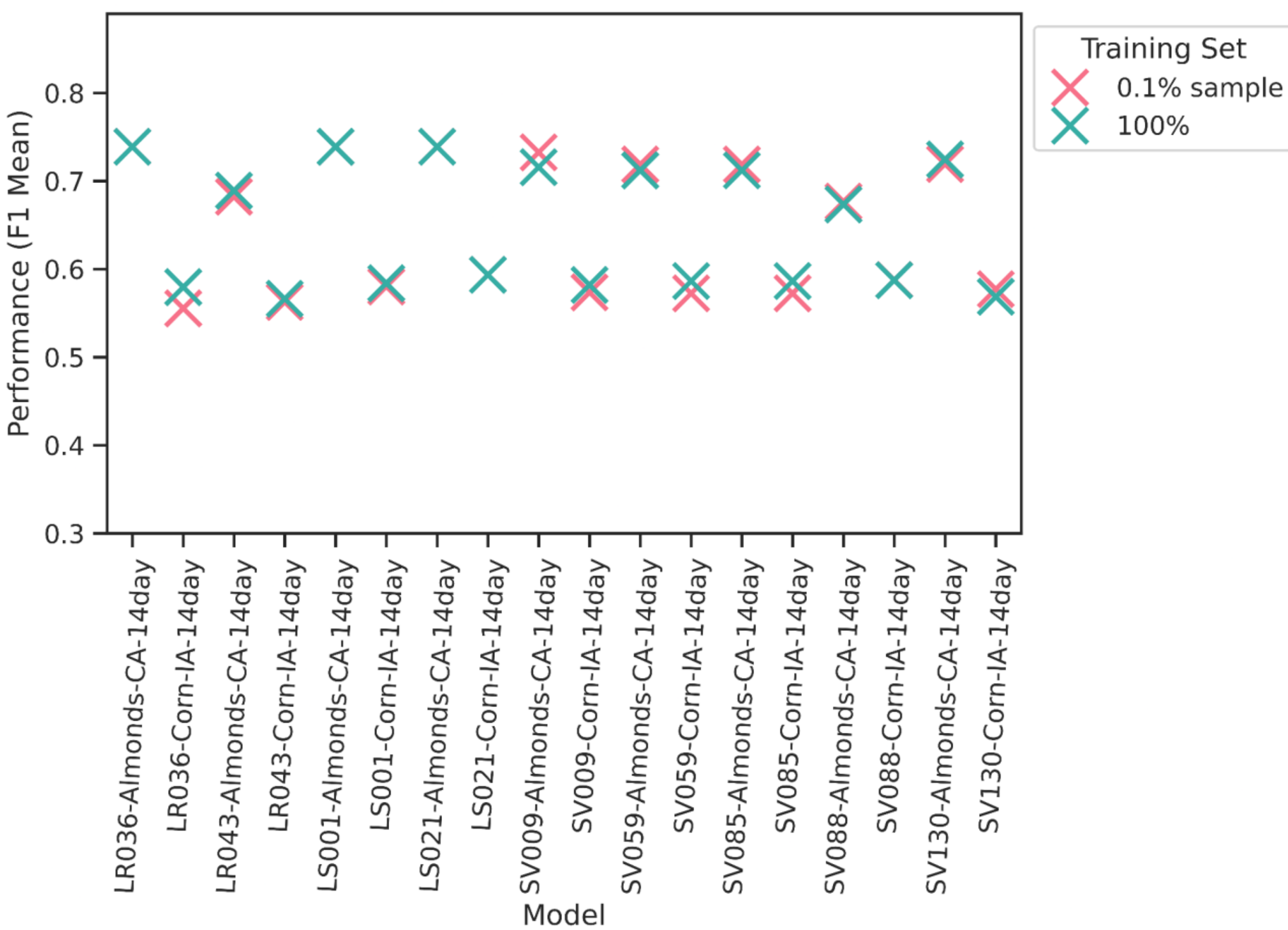

Figure 5. Performance (F1 mean) for selected algorithms trained on 0.1% sample as well as 100% of training data via bagging. There was no improvement when training on 100% data versus 0.1%. The only improving models were not the very best, so no new-best models were found.

*F. Case Study: Interannual patterns for California almonds by early June*

Comparing all 1582 models for almonds in California by early June reveals that among all these models and for every pair of years, F1 score in one validation year is positively correlated with F1 score in the other validation year, suggesting that in the universe of all models tested here, there are low performing models and high performing models, with no expected year-specific tradeoffs when switching from a low performing model to a higher performing model. Figure S10 shows pairwise plots of F1 scores in different validation years. However, comparing only the top 38 models for almonds in California by early June reveals that some pairs of validation years have negatively-correlated F1 scores. See Figure 6 and Table VII – pairs of years 2018 with 2019, 2018 with 2020, and 2018 with 2021 have negative correlations. This means that in the space of the *best* models, the almond mapping problem presents tradeoffs in choosing one model over another.

The distribution of almonds in the study area changed significantly in 2020, with otherwise incremental changes in almond distribution year to year. Table VIII shows prevalence of almonds on the diagonal, and year-to-year switching rate off-diagonal. The prevalence of almonds in 2020 was triple what it was in 2019, but within 2018-2019 the prevalence was similar, and within 2020-2022 it was similar as well. There was very low switching between 2018 and 2019, with rate of switching 0.002. Comparing either 2018 or 2019 to any of the 2020-2022 years shows a lot of switching due to the influx of new almond orchard acreage that occurred in 2020. There was low switching between 2020 and 2021 (rate of switching 0.012) and low switching between 2021 and 2022 (rate of switching 0.012). Therefore 2022 was even more different from 2020 (rate of switching 2020 to 2022 was 0.019).

Characterizing the feature space revealed that the most extreme differences in almond features year to year were in the crop rotation history features. The reflectance features of almonds were similar year to year; in any given period, the near-infrared band is negatively correlated with the other bands, and every other band was positively correlated with the others. Only slight differences are apparent for the first or second period in some years, likely due to weather conditions (Figure S3). The temporal patterns and spread of each band are also similar year to year for almonds, with only a slightly weaker blue and green reflectance in 2019, and a slightly wider interquartile range for blue, green, NIR and SWIR2 bands in 2020-2022 (Figures S4-S8). However, the crop rotation history patterns of almonds showed starker interannual differences, shown in Figure 7. The proportions in each box are the proportion of almond pixels that had history of being almonds. Each column thus shows a specific year's almonds' history profile, and their differences are clearly visible.

The case study of K-Nearest Neighbors hyperparameter interactions with almonds in different years revealed different optimal settings for years with different phenology-history profiles. To optimize performance for an unknown year, looking at mean F1 across years reveals that best performance comes from certain values of K and certain choices of metric, shown in Figure 8. K between 10 and 50 was the best, with Chebyshev the best distance metric, followed by Euclidean. Manhattan and Canberra metrics produced worse performance, and for these the 14-day scheme is always better than 5-day scheme, but for the Euclidean and Chebyshev metrics, the effect of scheme is unclear. Distance weighting was also analyzed, and among the best-performing KNN models it is unclear whether distance-weighting (not pictured) affects performance. However, individual years produced best performance in different hyperparameters, shown in Figures S11-S15. In 2018, Manhattan distance and K between 10 and 50 was best. In 2019, Manhattan, Chebyshev, or Euclidean distance with K between 8 and 80 was best, or Euclidean distance with K between 200 and 500. In 2020, it was Manhattan distance with K between 6 and 12, or Chebyshev distance with K between 100 and 500, or Euclidean distance with K between 30 and 200. In 2021, the best was Chebyshev distance with K between 10 and 100 or K around 800. Finally in 2022 the best was Manhattan with K between 10 and 80, or Euclidean with K between 6 and 400, or Chebyshev with K between 6 and 300. In every year, the 14-day scheme is better for Manhattan and Canberra distance, but for Euclidean and Chebyshev distance the effect of scheme is unclear. In general Manhattan and Canberra distance reached optimal performance in conjunction with smaller K, while Euclidean and Chebyshev reached optimal performance at a variety of values of K. Overall, the best hyperparameter settings vary substantially by validation year.

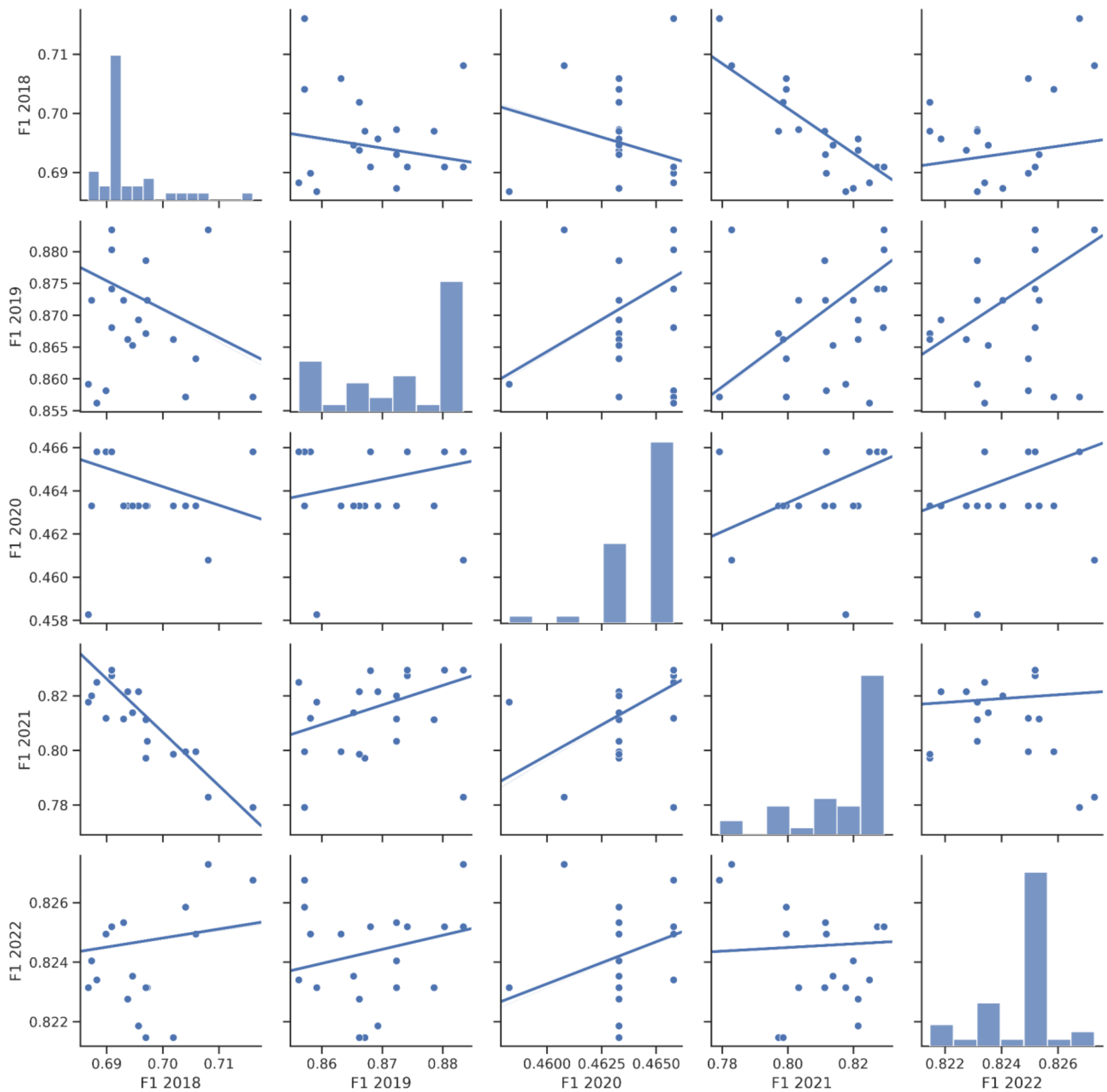

Figure 6. Off-diagonal: Pairwise F1 scores across validation years for the top 38 (within 0.01 mean F1 score of best) Almonds-California-June models. The following pairs of years have negative correlations: 2018 with 2019, 2018 with 2020, and 2018 with 2021. On-diagonal: Histogram of F1 scores for each validation year for the top 38 Almonds-California-June models.

Table VII

Pearson correlation coefficients for pairwise F1 scores across validation years for the top 38 (within 0.01 mean F1 score of best) Almonds-California-June models. The off-diagonal values correspond to the trend lines in Figure 9. The following pairs of years have negative correlations: 2018 with 2019, 2018 with 2020, and 2018 with 2021.

| | *2018* | *2019* | *2020* | *2021* | *2022* |
|---|---|---|---|---|---|
| *2018* | 1 | -0.27 | -0.31 | -0.86 | 0.14 |
| *2019* | -0.27 | 1 | 0.34 | 0.52 | 0.37 |
| *2020* | -0.31 | 0.34 | 1 | 0.55 | 0.37 |
| *2021* | -0.86 | 0.52 | 0.55 | 1 | 0.07 |
| *2022* | 0.14 | 0.37 | 0.37 | 0.07 | 1 |

Table VIII

DIAGONAL: PREVALENCE OF ALMONDS IN EACH YEAR IN THE SACRAMENTO-AREA STUDY REGION. OFF-DIAGONAL: PREVALENCE OF SWITCHING BETWEEN THE TWO YEARS. SWITCHING INCLUDES ALMONDS TO NON-ALMONDS AS WELL AS NON-ALMONDS TO ALMONDS.

| | *2018* | *2019* | *2020* | *2021* | *2022* |
|---|---|---|---|---|---|
| *2018* | 0.010 | 0.002 | 0.026 | 0.025 | 0.026 |
| *2019* | 0.002 | 0.011 | 0.024 | 0.023 | 0.025 |
| *2020* | 0.026 | 0.024 | 0.035 | 0.012 | 0.019 |
| *2021* | 0.025 | 0.023 | 0.012 | 0.034 | 0.012 |
| *2022* | 0.026 | 0.025 | 0.019 | 0.012 | 0.035 |

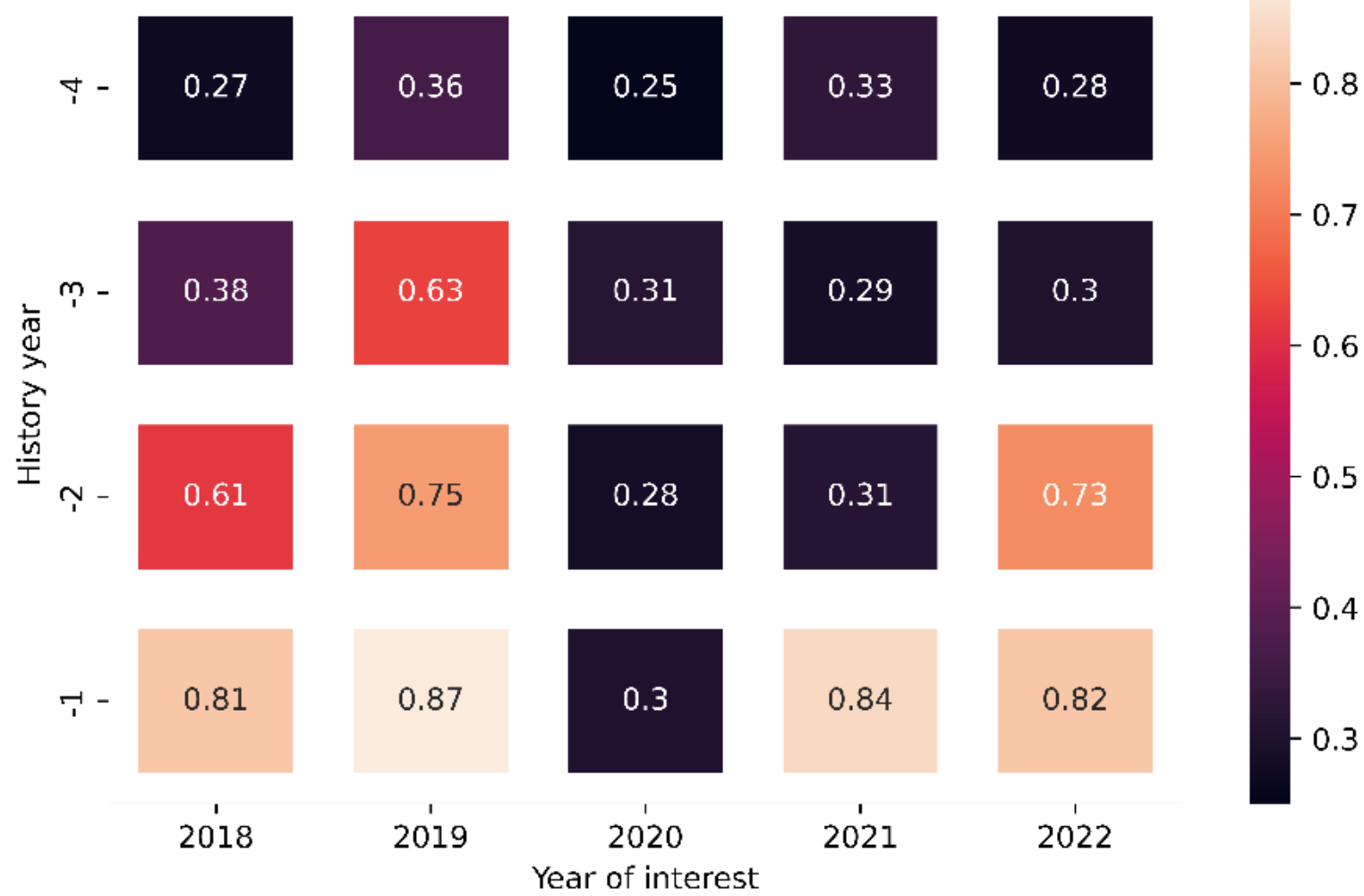

Figure 7. Crop rotation history of almond pixels. Each column shows a history profile for almonds in a specific year of interest.

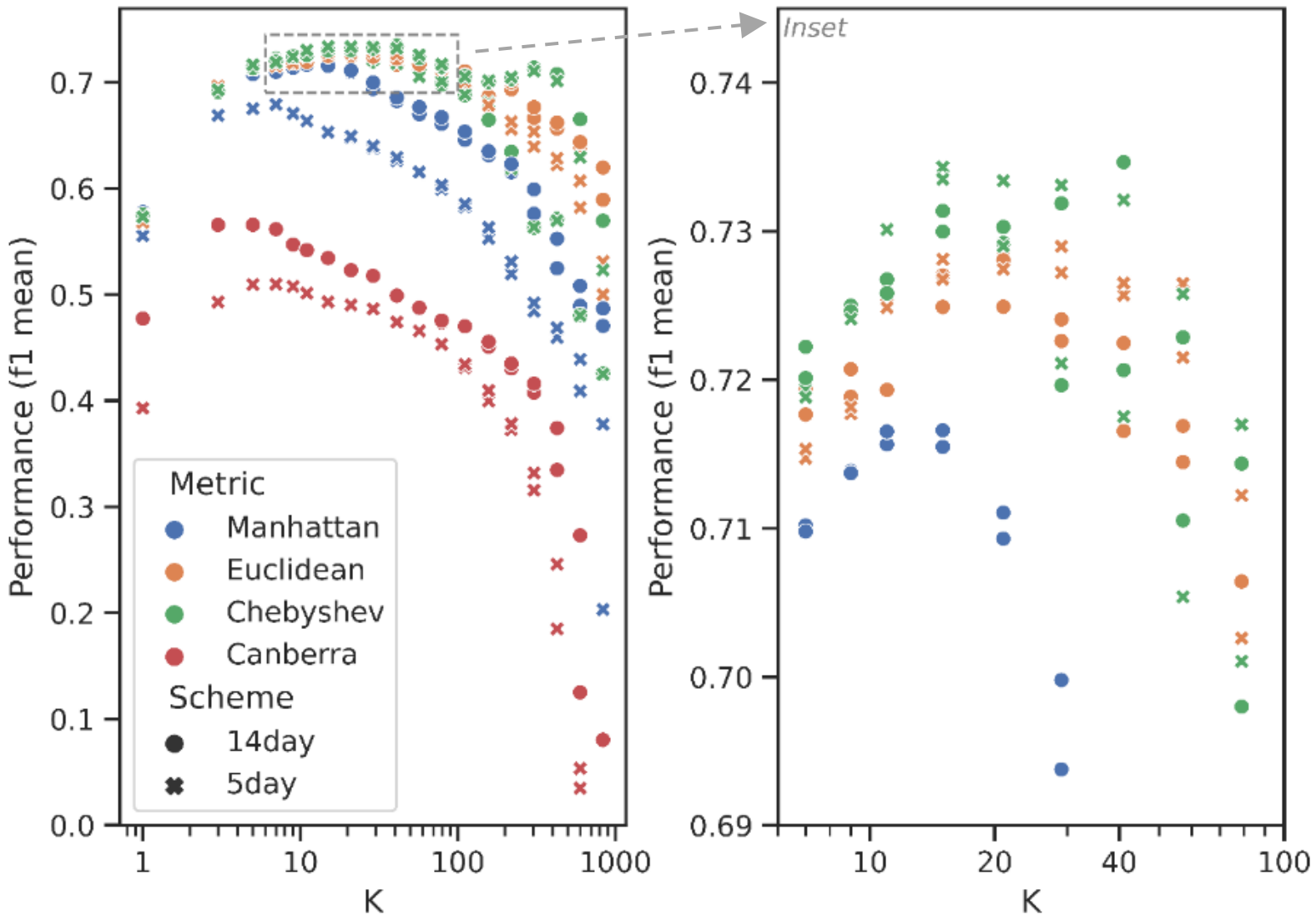

Figure 8. Performance of K Nearest Neighbors algorithms based on choice of hyperparameters: distance metric, compositing scheme, and K number of neighbors.

## IV. Discussion

A crop type mapping methodology was developed that can create accurate maps in unseen years by early June of the unseen year. This study quantified the uncertainty around accuracy, so that farmers, policymakers, and emergency managers have an honest idea of how accurate a new year's map might be. Finally, interannual variability in crop features and distribution means different algorithms perform best in different conditions, suggesting an opportunity to further improve performance using ensembles or ancillary data.

### *A. Best algorithms by crop, region, in-season date, and training size*

The best model by mean F1 score for California almonds by early June achieved a mean F1 score of 0.74. The mean precision across years was 0.73, meaning 73% of the early-June-mapped almond pixels are actually almonds in the CDL.

The mean recall across years was 0.83, meaning 83% of the almond pixels in the CDL appear as almonds in the early-June map.

The best model by mean F1 score for Iowa corn by early June achieved a mean F1 score of 0.59. The mean precision across years was 0.60, meaning 60% of the early-June-mapped corn pixels are actually corn in the CDL. The mean recall across years was 0.62, meaning 62% of the corn pixels in the CDL appear as corn in the early-June map.

Better F1 scores were found for California almonds than for Iowa corn. One explanation is that Iowa cropland is dominated by corn and soybeans, with soybeans having similar phenology to corn and thus becoming a frequent confusion crop. On the other hand, California almonds have pistachios and walnuts nearby with similar-enough phenology to become confusion crops, but there are fewer pixels of pistachios and walnuts and thus confusion happens less frequently. Another relevant dynamic is that many Iowa farms rotate between corn and soybeans interannually. While crop rotation history features help the algorithm learn the rotation patterns (for example, a corn-soybean-corn-soybean pattern for the last four years suggests corn in the year of interest), farmers also frequently rotate crops in unpredictable ways (for example, choosing to plant soybeans two years in a row after a history of switching every year). On the other hand, almonds grow in orchards, where farmers keep the same orchard trees growing many years in a row. Therefore, crop rotation history information is especially helpful for almonds – for example, a pixel with almonds-almonds-almonds-almonds for the previous four years is extremely likely to be almonds in the year of interest. For this not to be the case means the farmer either ripped out the almond orchard or let it decay severely.

The California almonds best model showed substantial interannual variation in performance, as indicated by the year-to-year F1 standard deviation of 0.15. The five years (2018-2022) comprising that standard deviation had scores of 0.69, 0.88, 0.47, 0.83, 0.83 respectively, and all the top 38 models had a similar distribution. The substantial spread of F1 scores across these five years suggests that year-to-year variation in crop phenology and distribution, due to weather and farming decisions, play a significant role in the accuracy of a crop type map in an unseen year. Therefore, the authors encourage other researchers to report performance metrics on multiple validation years, rather than on just one year as has been the norm.

The fact that early June models achieved F1 scores on par with end-of-year models was somewhat surprising. The explanation is that crop rotation history plays a significant role in conjunction with HLS reflectance. Both contribute to making the best map, but it appears that the early-season reflectance information was enough reflectance information. Some machine learning algorithms also have challenges with higher-dimensional data, and in this case more reflectance information means more feature dimensions. Further research could identify which features contribute most to the model's crop classification, and analyzing this information for the end-of-season reflectance dataset may reveal why the late-season features cannot provide any performance gains.

Training on 0.1% of data was just as good as 100%. The geospatial reason is that the 0.1% sample captured most corn fields (or almond orchards), which contain most of the phenological variation across corn (or across almonds) in a given year. Likewise, that 0.1% sample captured most kinds of non-corn (or non-almond) land cover, training the machine learning model to distinguish the class of interest from the negative class. In other words, the 100% set contains pixels that are spatially near the 0.1% sample pixels, so nearly every point in the 100% set is well-represented by some point in the 0.1% set – therefore, the 0.1% sample is enough to train the machine learning model effectively.

### *B. Caveats: reprojected CDL and five validation years*

The Cropland Data Layer product is developed by USDA via machine learning on remote sensing data, using on the ground surveys from the Farm Service Agency as ground truth. Precision and recall of the CDL are published by state, year, and crop. For corn in Iowa, both precision and recall are roughly 0.96 for every year 2018-2022. For almonds in California, both precision and recall are roughly 0.90 for every year 2018-2022. The performance scores of the in-season crop maps in this study were reported relative to the CDL. CDL serves as labeled training data for our models, and thus our study inherits the potential errors and limitations of the CDL products.

The CDL was reprojected and resampled from Albers Equal Area projection to UTM projection consistent with the HLS grid using a nearest-neighbor resampling approach. Reprojection and resampling may introduce minor errors in the datasets.

The year-wise cross validation scheme employed in our study provides a robust estimate of model performance for model application to an unseen year, because it uses five different unseen years. This performance estimate thus assumes that the future range of farming conditions and crop spectral response is similar to what was seen during these five years. As shown in this study, using five unseen years provides a much better estimate than using one unseen year, but future unseen extreme events would cause even this estimate to be incomplete.

### *C. Best practices for optimizing prediction performance*

Analysis of results from the study provides insights into the strengths and weaknesses of various modeling choices. Regarding compositing schemes, the datasets represented by 5-day periods usually were generally on par with those represented by 14-day periods, although 14-day compositing periods provided the very best classification performance. Linear and nonlinear SVMs perform well for both almonds in California and corn in Iowa, and future research could focus on further refining SVM-based algorithms. Hyperparameter search was critical, since each algorithm's optimal hyperparameters vary depending on crop type, in-season cutoff, and compositing scheme. In this study, multiple algorithms had a hyperparameterization that demonstrated similarly high performance in this study's robust evaluation. However, a more exhaustive hyperparameter search may yet discover better performing models for in-season corn and almond mapping, or for other specific applications.

### *D. Connecting hyperparameters with interannual variation*

Looking closely at California almonds early-June models, the interannual variation of crop phenology and distribution caused different validation years to have different performance and different optimal hyperparameterizations. The best California early-June model had scores of 0.69, 0.88, 0.47, 0.83, 0.83 respectively, for years 2018-2022, and all the other top 38 - the ones within 0.01 mean F1 score of best - were no more than 0.03 better in every year. With such similar interannual performance profiles, stakeholders should seek to understand interannual variation in map accuracy rather than seek performance gains from switching among the top 38 models.

Characterizing the feature space and the distributions of almonds provides explanations for why 2020 was such a difficult year for in-season crop mapping. Because almond orchard area tripled from 2019 to 2020, out of the 2020 almonds only 30% of them had previous-year history of being almonds, compared with the typical 80% (Figure 7), causing the machine learning model to perform worse in this circumstance. In real world operations, drastic changes like a tripling of almond acreage, or mass removal of almond orchards due to drought, would be known and end-users would know not to trust the model. This opens up opportunities for research in training new models to anticipate unusual years.

Seeing that different K Nearest Neighbors hyperparameters were best in different years implies that certain types of years lend themselves to certain algorithm settings. Since ancillary information, like early season weather as well as economic conditions incentivizing farmers to plant certain crops, implies specific algorithm settings will perform better, incorporating these could improve in-season crop mapping performance.

In this study, the task of connecting hyperparameters to types of years was only just begun, with analysis of K Nearest Neighbors revealing some insights. The 2021 validation year (Figure S21) was the only one in which the Manhattan distance metric was much worse than the Chebyshev distance metric, and we hypothesize this is due to the crop rotation history features. If multiple or all crop rotation history years do not match, then the Chebyshev metric cares just as much as if only one year did not match; however, for the Manhattan distance metric each additional nonmatching year matters just as much in deciding that other datapoint is a non-neighbor. Hence, a theory emerges for why Chebyshev outperformed Manhattan in 2021. While the reflectance features of almonds were relatively similar in all years, the crop rotation history profiles of 2021 almonds are unique, with a high rate of previous-year almonds, but a low rate of almonds two to four years prior. Chebyshev distance with K between 10 and 100 allows the algorithm to latch onto those few training datapoints that exactly match the dominant 2021 crop rotation history pattern, while Manhattan distance accepts some training datapoints with nonmatching crop rotation history as near neighbors. Thus, the unique crop rotation history of 2021 almonds, in a year when the phenology was very similar to training data, created the right conditions for Chebyshev distance to outperform Manhattan distance. This analysis establishes best practice to clarify the relationship between hyperparameter choices (e.g., Chebyshev versus Manhattan distance metric) and the situations they perform best in; future studies can extend it beyond K Nearest Neighbors to other algorithms.

## V. Conclusion

This study compared ten machine learning algorithms for mapping crop types by early June of an unseen year, and provided evaluations across five unseen years to help practitioners trust the model in the face of uncertainty. The algorithms were compared for accuracy and interannual variability in classifying corn in Iowa and almonds in California, based on Harmonized Landsat Sentinel surface reflectance and Cropland Data Layer crop rotation history. Support Vector Machines exhibited the best all-around performance by early June, with mean F1 scores of 0.74 for almonds in California and 0.59 for corn in Iowa, so this algorithm is a good choice for future work. However, hyperparameter optimization is necessary to achieve the best performance, and interannual variation means that some algorithms and settings are more reliable than others. This study quantified the interannual uncertainty to provide a robust expectation of performance in an unseen year, which is the real-world use case for in-season crop mapping. This work helps farmers, emergency managers, and policymakers make food security decisions as natural hazards like floods and insects become more common due to climate change and threaten crops more frequently.

In the future, ensembles and ancillary data will be investigated to further improve the accuracy of in-season crop maps. The scope of the mapping can also extend to multiclass, mapping all common crops in a given region, or extend across the conterminous United States. Given more labeled training data, this methodology can extend globally. A final vision is an easy-to-use online tool for stakeholders, such as farmers, scientists, policymakers, and emergency managers.

## Acknowledgment

The authors thank Shashank Konduri, Nathan Beningson, Hanna Renedo, Tianqi Zhang, Morgan Steckler, Robyn Anderson, Jack Watson, and Raviraj Dave for their feedback during this study.

**August Posch** received a B.A. in mathematics from Bowdoin College, Brunswick, ME, USA in 2018, and M.S. in data science from Roux Institute and Khoury College of Computer Sciences, Northeastern University, Portland, ME, USA in 2022.

He currently works as a Data Scientist and post-M.S. Research Scientist with the AI for Climate and Sustainability group and Sustainability and Data Sciences Laboratory within Northeastern University, Boston, MA, USA. His research interests include remote sensing, machine learning, uncertainty quantification, domain adaptation, climate modeling, extreme weather, downscaling, and networked infrastructure resilience.

**Jitendra Kumar** (Senior Member, IEEE), photograph and biography not available at the time of submission.

**Forrest M. Hoffman** (Senior Member, IEEE), photograph and biography not available at the time of submission.

**Auroop R. Ganguly** (Senior Member, IEEE), photograph and biography not available at the time of submission.

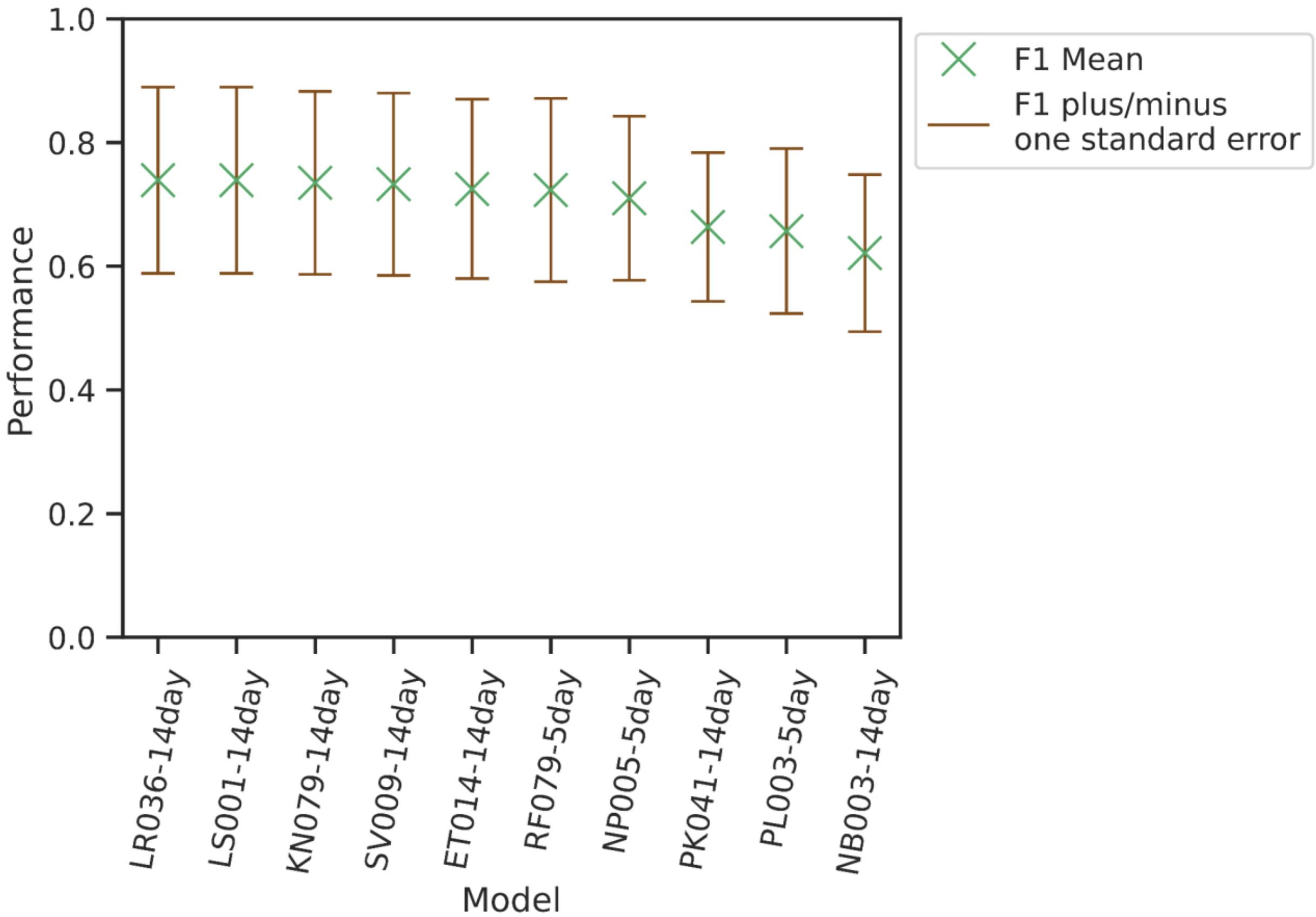

Figure S1. Performance with standard error for the best configuration of each architecture for almonds in California by early June.

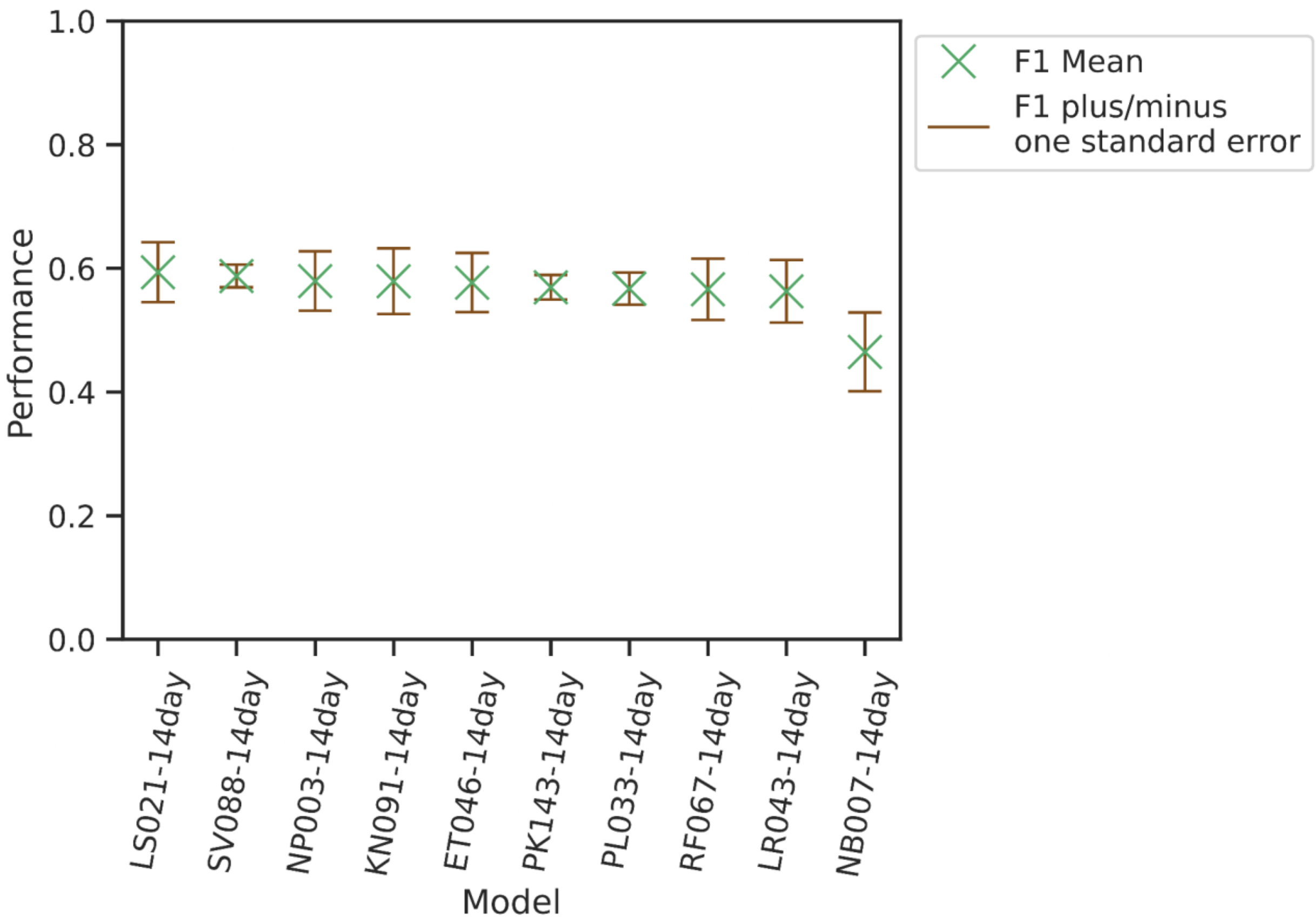

Figure S2. Performance with standard error for the best configuration of each architecture for corn in Iowa by early June.

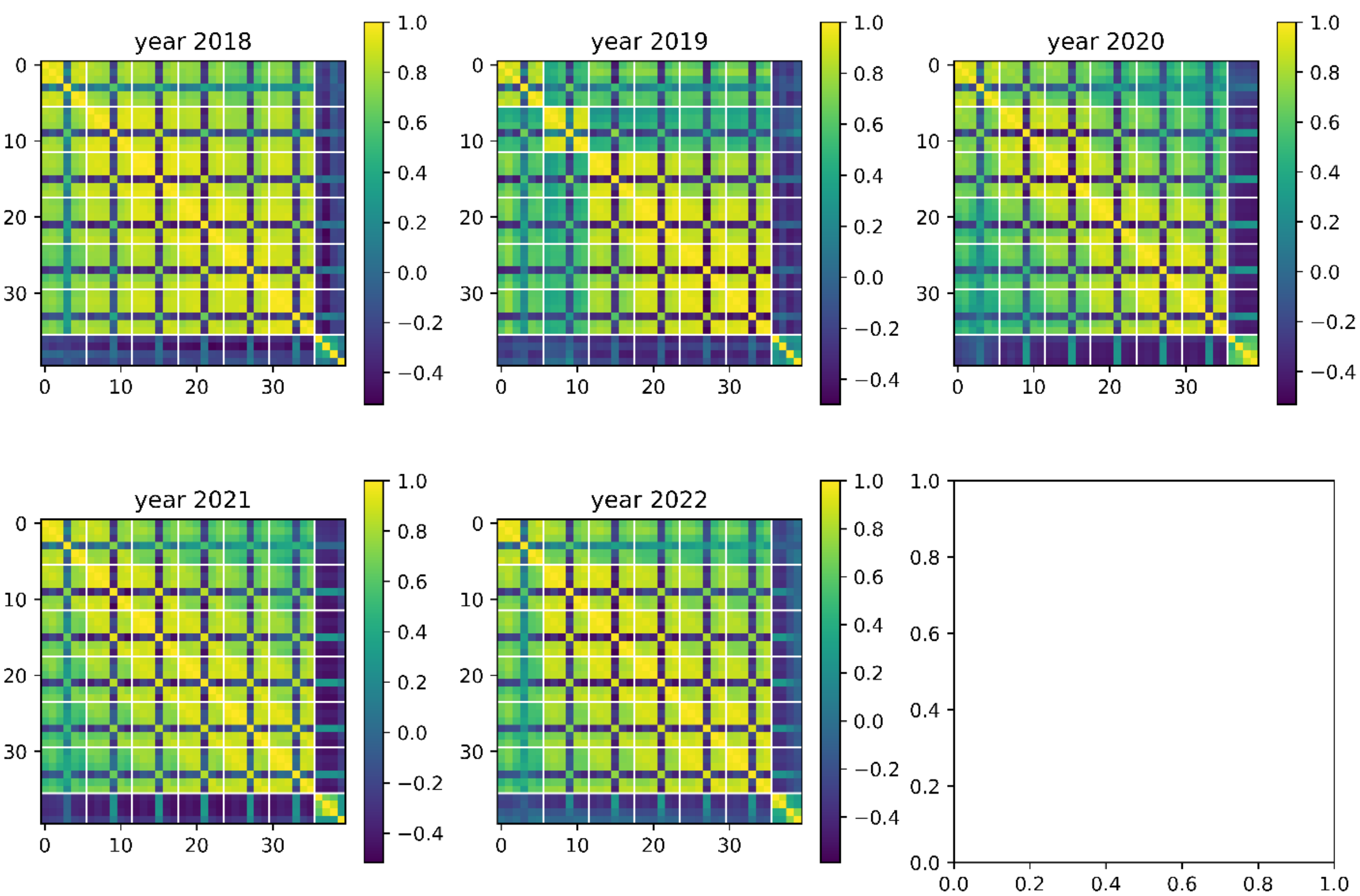

Figure S3. Feature space correlations for almonds in each year. The 40 features shown include sets of 6 bands (blue, green, red, NIR, SWIR1, SWIR2) for each of 6 compositing periods, with compositing periods delineated by white separation lines. The last four features correspond to crop rotation history.

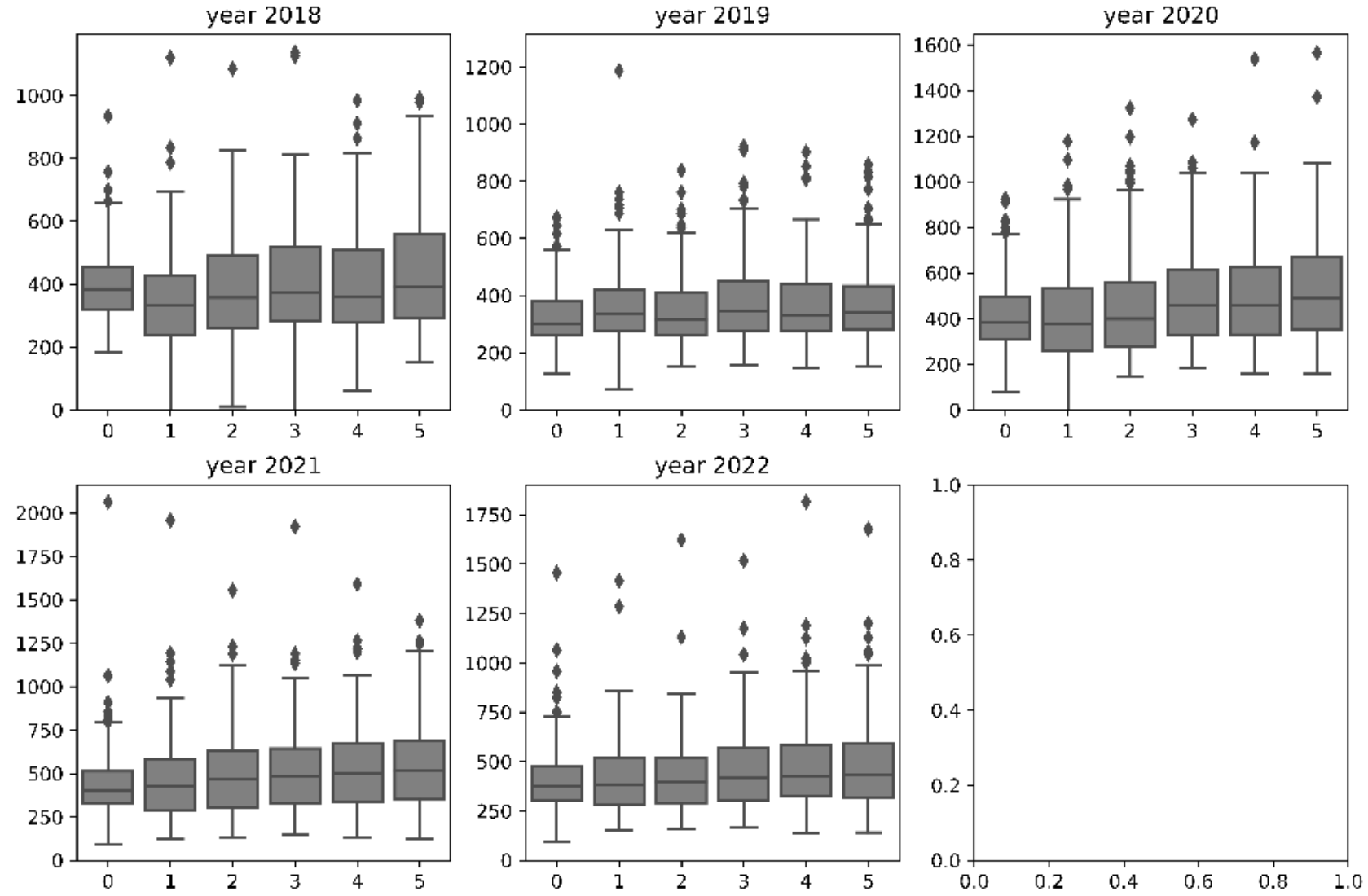

Figure S4. Blue reflectance by year by period. Here, period 0 means days 1-90, period 1 means days 91-104, period 2 means days 105-118, period 3 means days 119-132, period 4 means days 133-146, and period 5 means days 147-160. Reflectance scale has a factor of $10^5$, so 1000 on the plot means 10% of the sun's light was reflected back to the satellite.

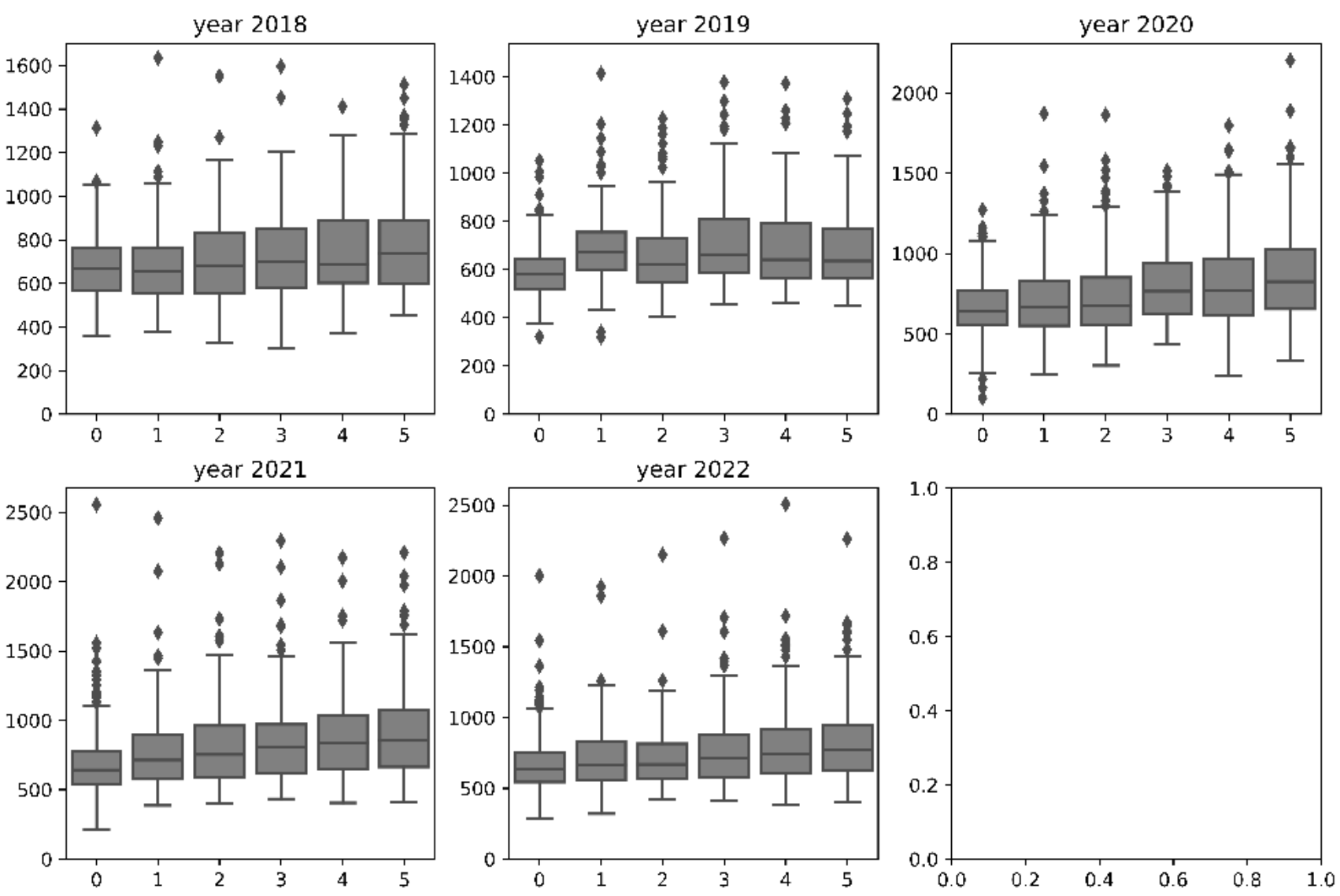

Figure S5. Green reflectance by year by period. Here, period 0 means days 1-90, period 1 means days 91-104, period 2 means days 105-118, period 3 means days 119-132, period 4 means days 133-146, and period 5 means days 147-160. Reflectance scale has a factor of $10^5$, so 1000 on the plot means 10% of the sun's light was reflected back to the satellite.

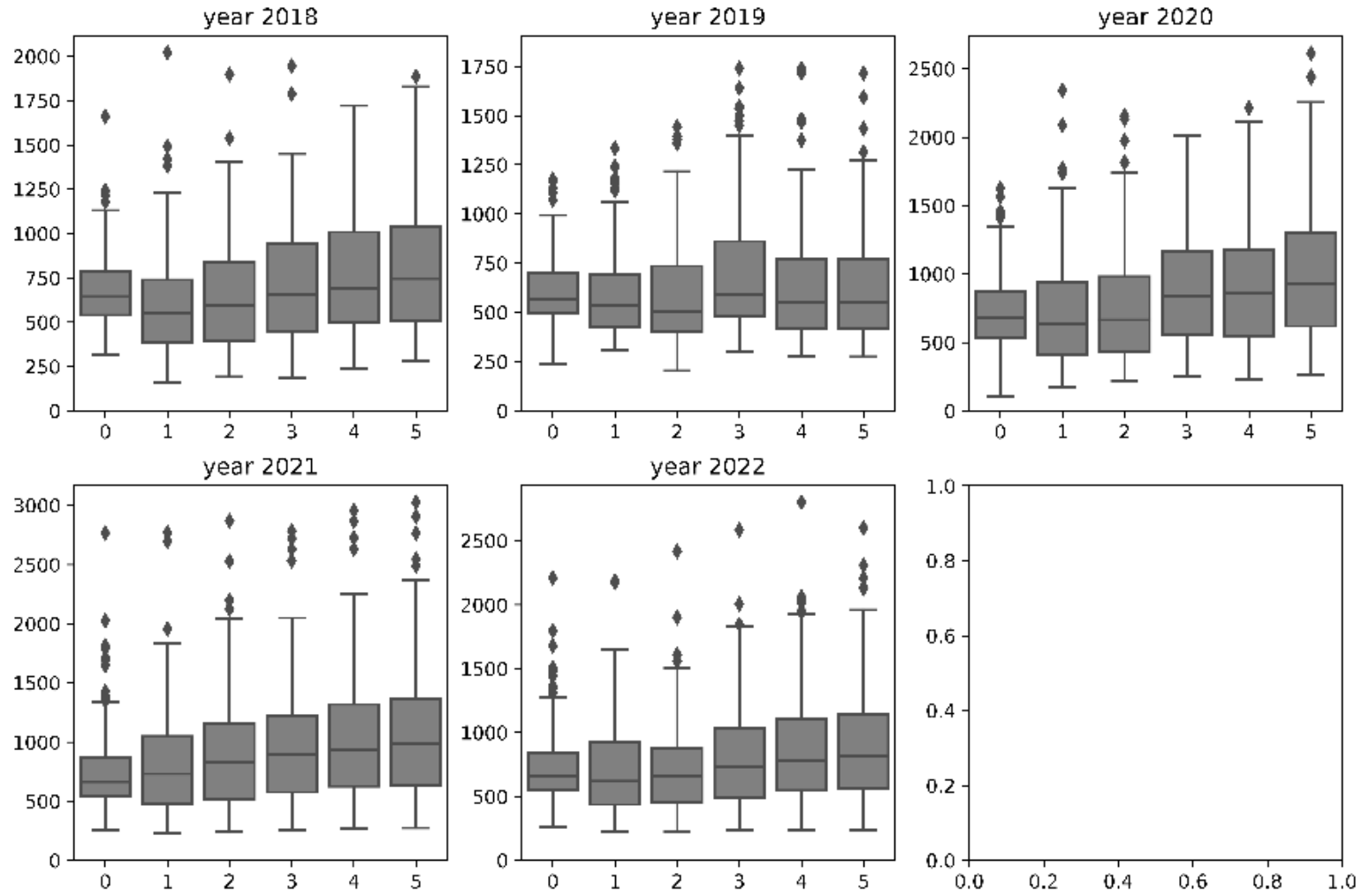

Figure S6. Red reflectance by year by period. Here, period 0 means days 1-90, period 1 means days 91-104, period 2 means days 105-118, period 3 means days 119-132, period 4 means days 133-146, and period 5 means days 147-

160. Reflectance scale has a factor of $10^5$, so 1000 on the plot means 10% of the sun's light was reflected back to the satellite.

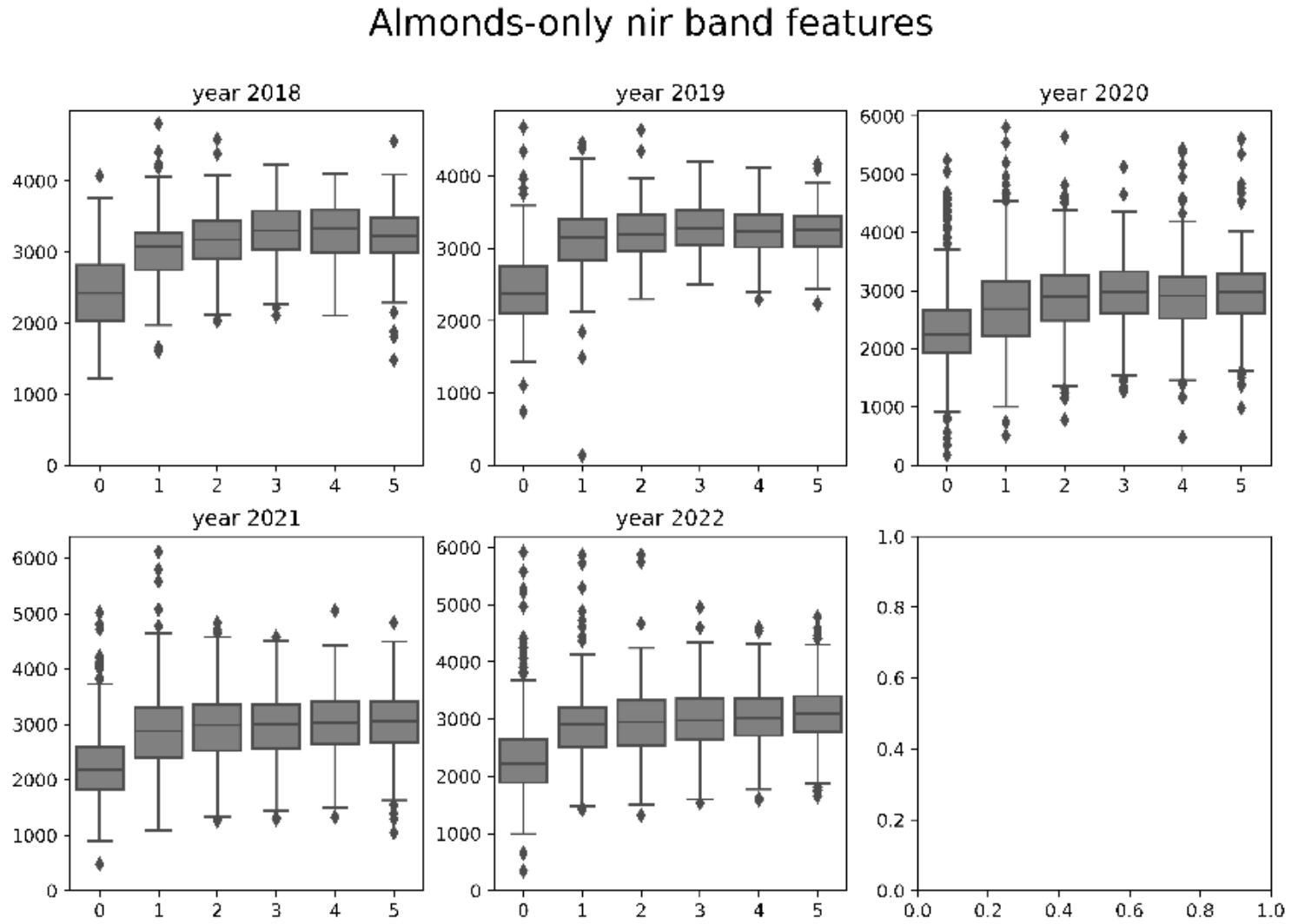

Figure S7. Near-infrared reflectance by year by period. Here, period 0 means days 1-90, period 1 means days 91-104, period 2 means days 105-118, period 3 means days 119-132, period 4 means days 133-146, and period 5 means days 147-160. Reflectance scale has a factor of $10^5$, so 1000 on the plot means 10% of the sun's light was reflected back to the satellite.

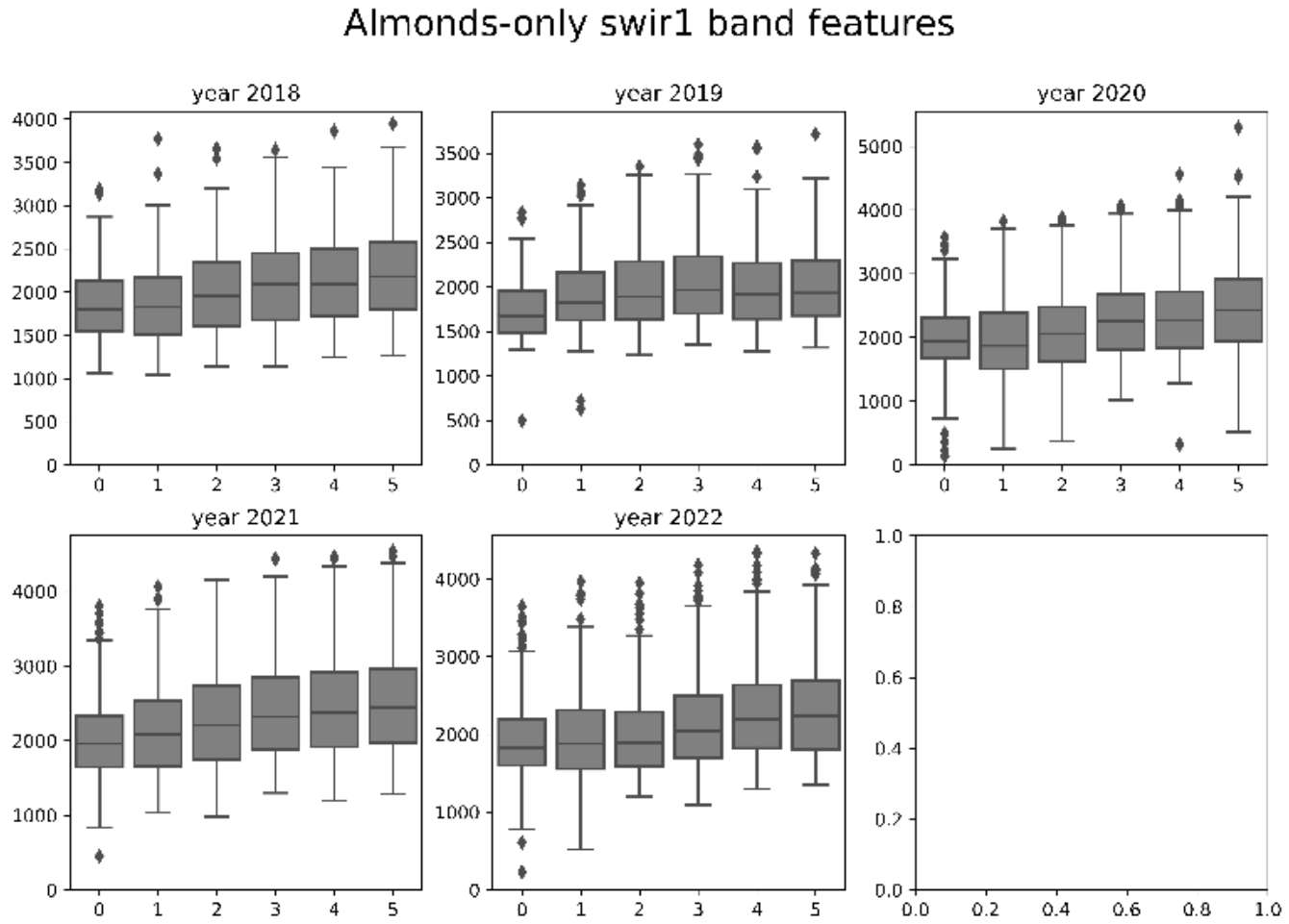

Figure S8. Shortwave infrared 1 reflectance by year by period. Here, period 0 means days 1-90, period 1 means days 91-104, period 2 means days 105-118, period 3 means days 119-132, period 4 means days 133-146, and period 5 means days 147-160. Reflectance scale has a factor of $10^5$, so 1000 on the plot means 10% of the sun's light was reflected back to the satellite.

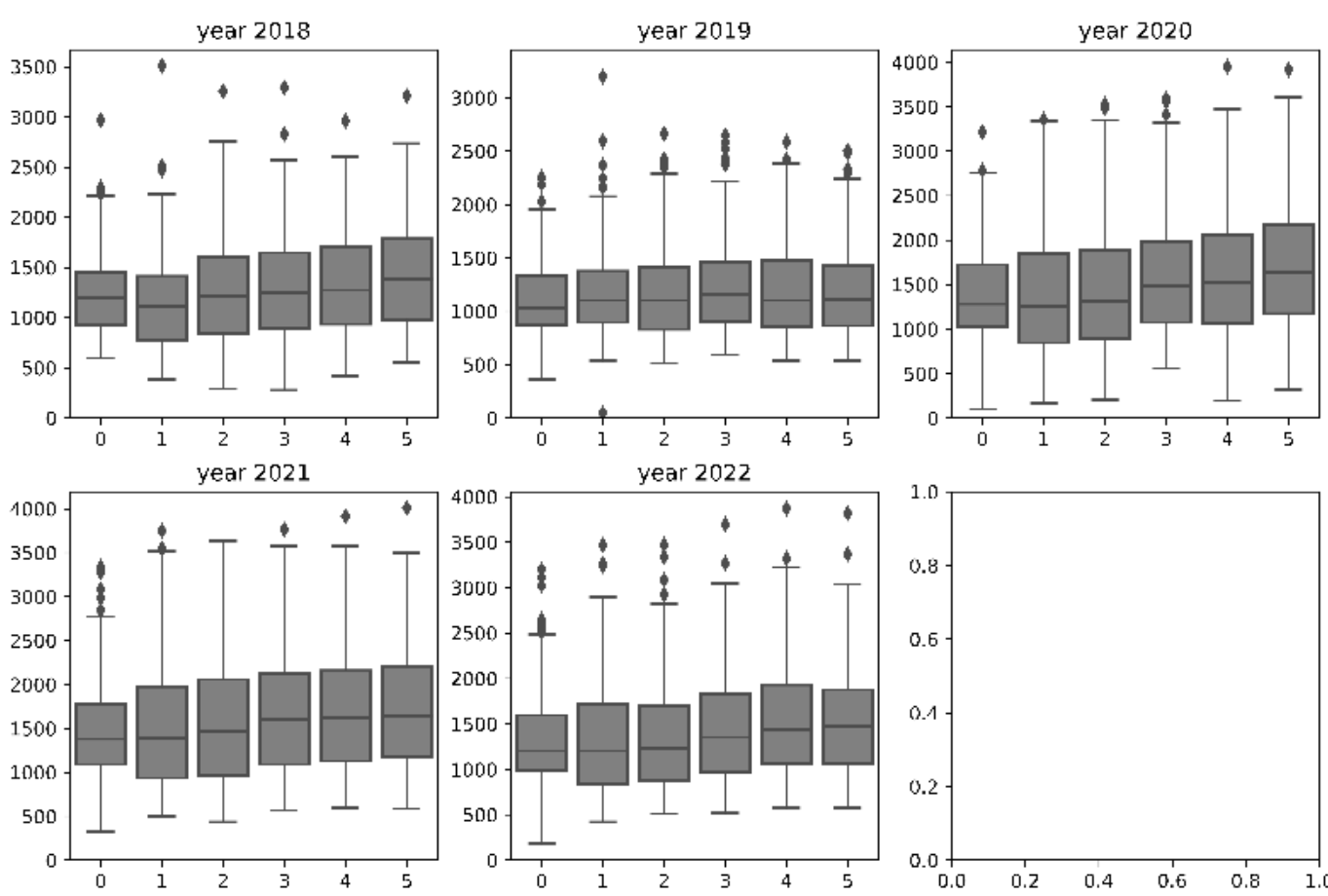

Figure S9. Shortwave infrared 2 reflectance by year by period. Here, period 0 means days 1-90, period 1 means days 91-104, period 2 means days 105-118, period 3 means days 119-132, period 4 means days 133-146, and period 5 means days 147-160. Reflectance scale has a factor of $10^5$, so 1000 on the plot means 10% of the sun's light was reflected back to the satellite.

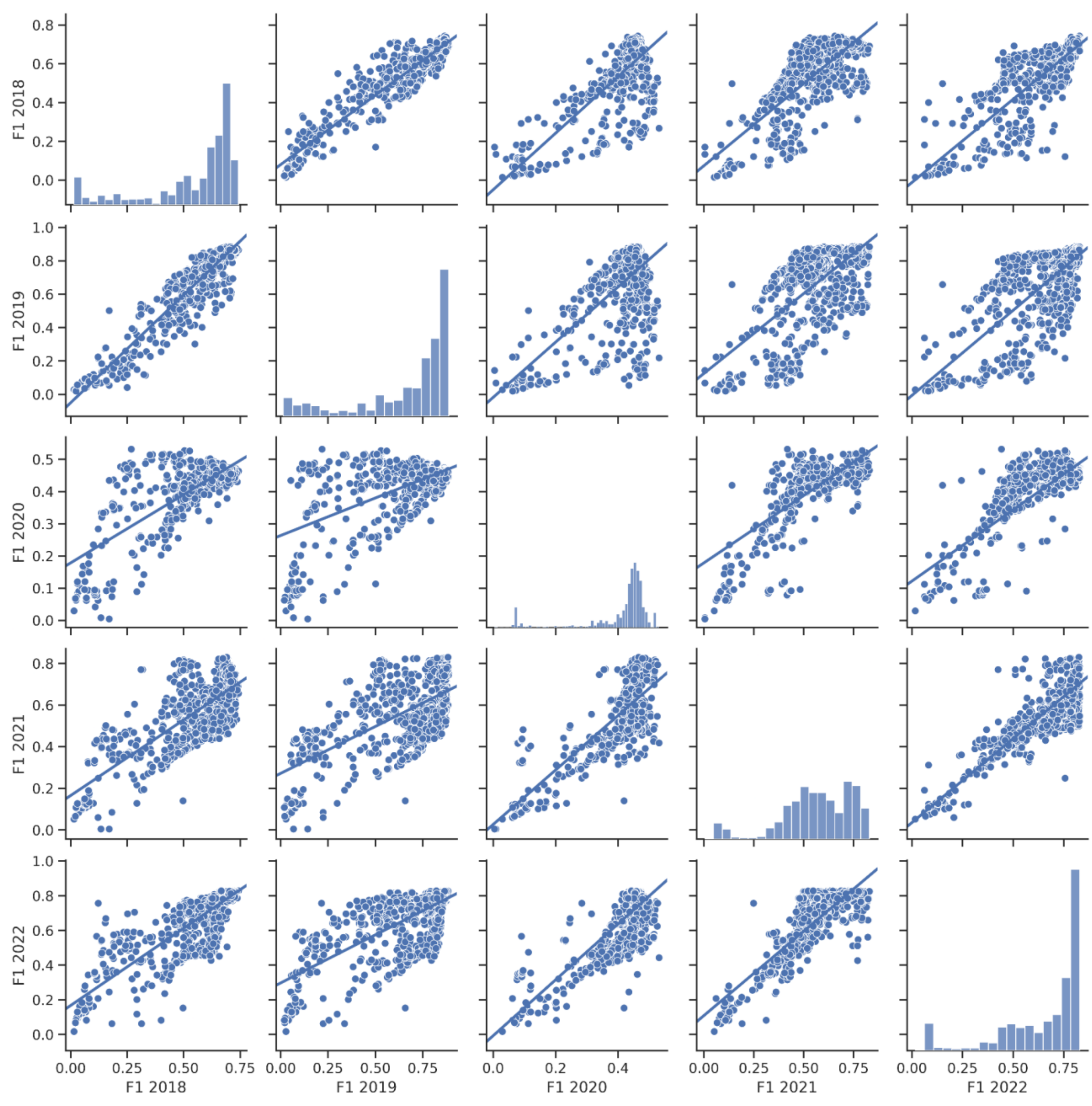

Figure S17. Off-diagonal: Pairwise f1 scores across validation years for all 1594 Almonds-California-June models. All pairs of years have positive correlations. On-diagonal: Histogram of f1 scores for each validation year for all 1594 Almonds-California-June models.

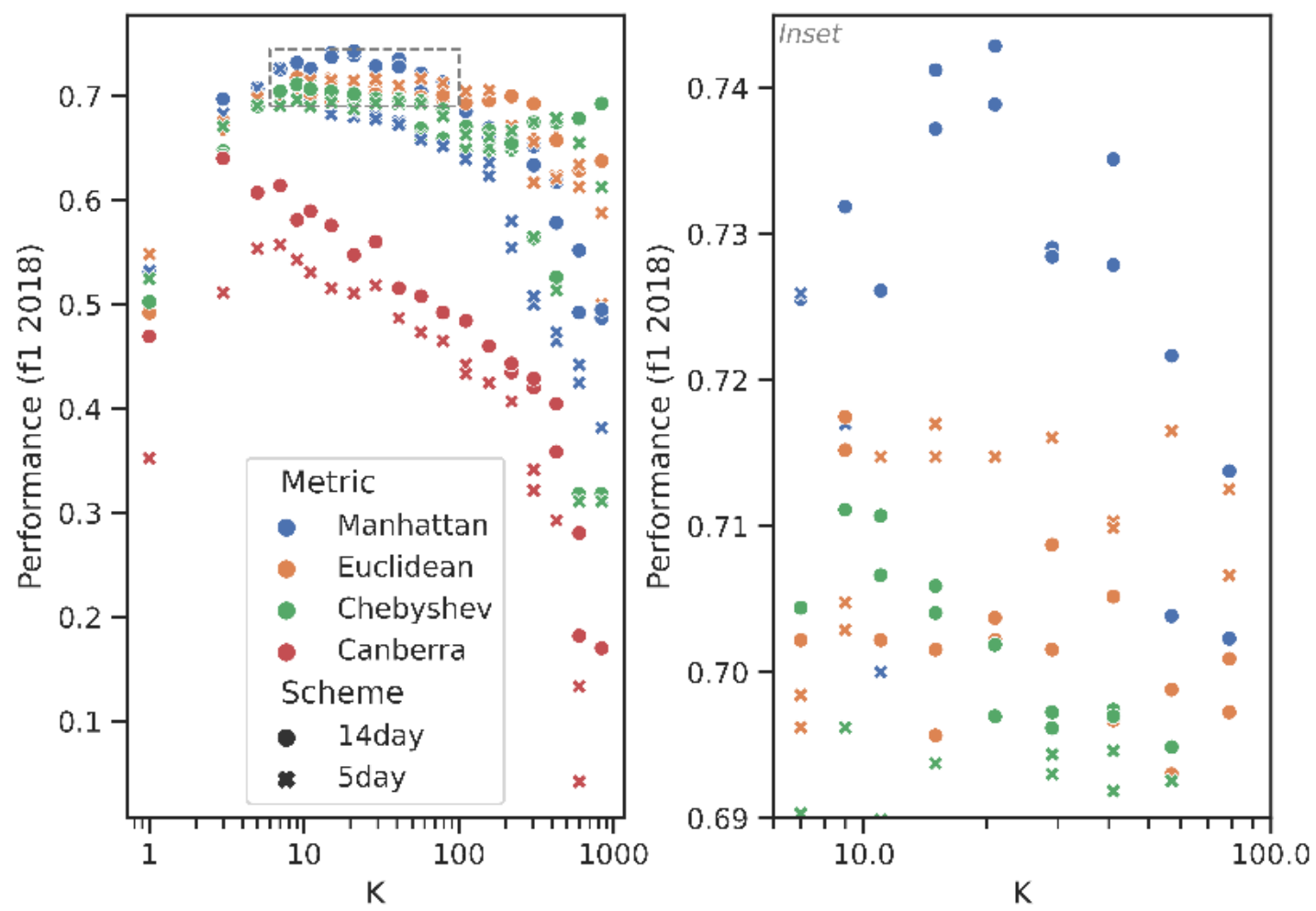

Figure S11. K-nearest Neighbors hyperparameter settings and performance for validation year 2018.

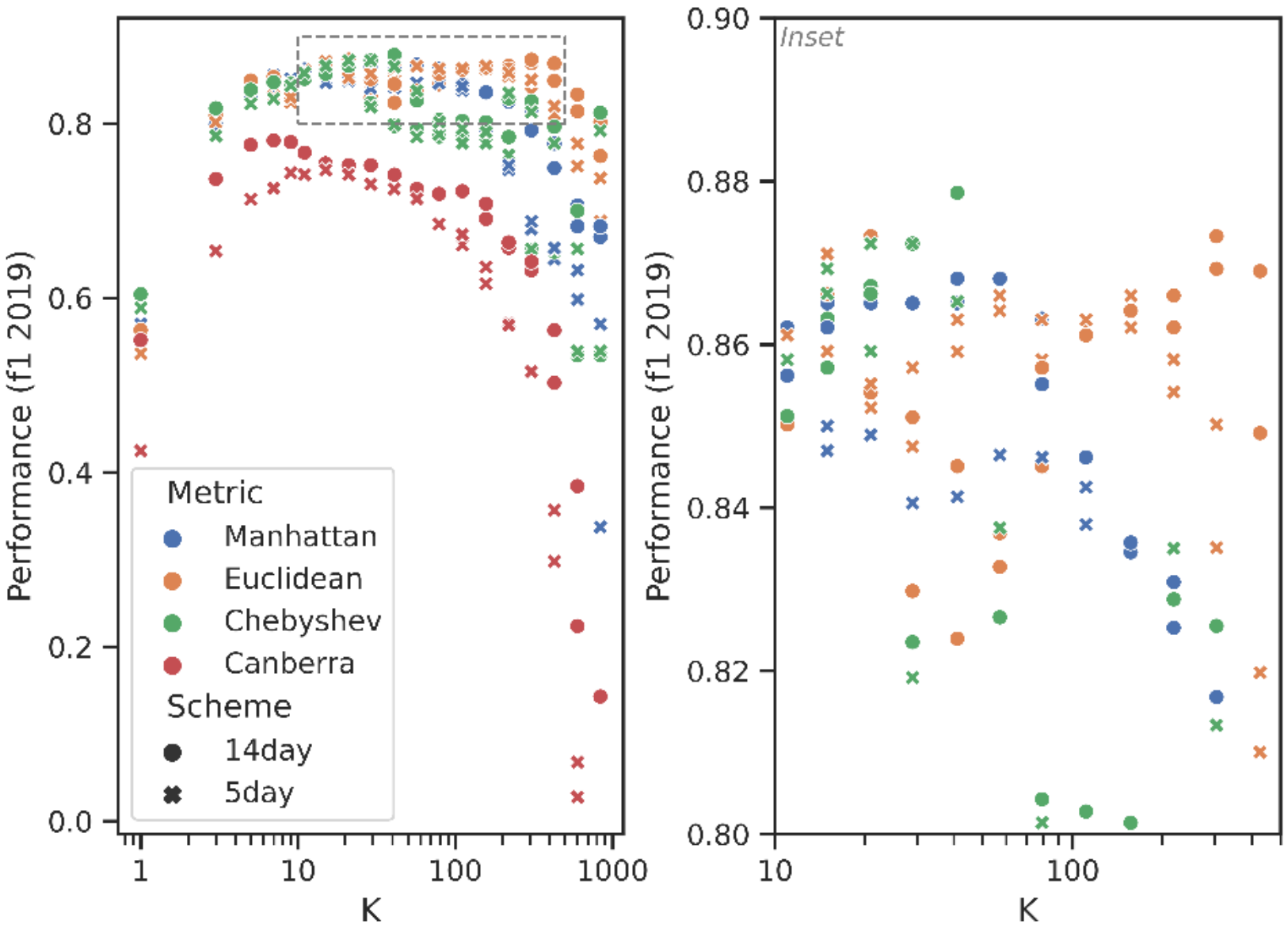

Figure S12. K-nearest Neighbors hyperparameter settings and performance for validation year 2019.

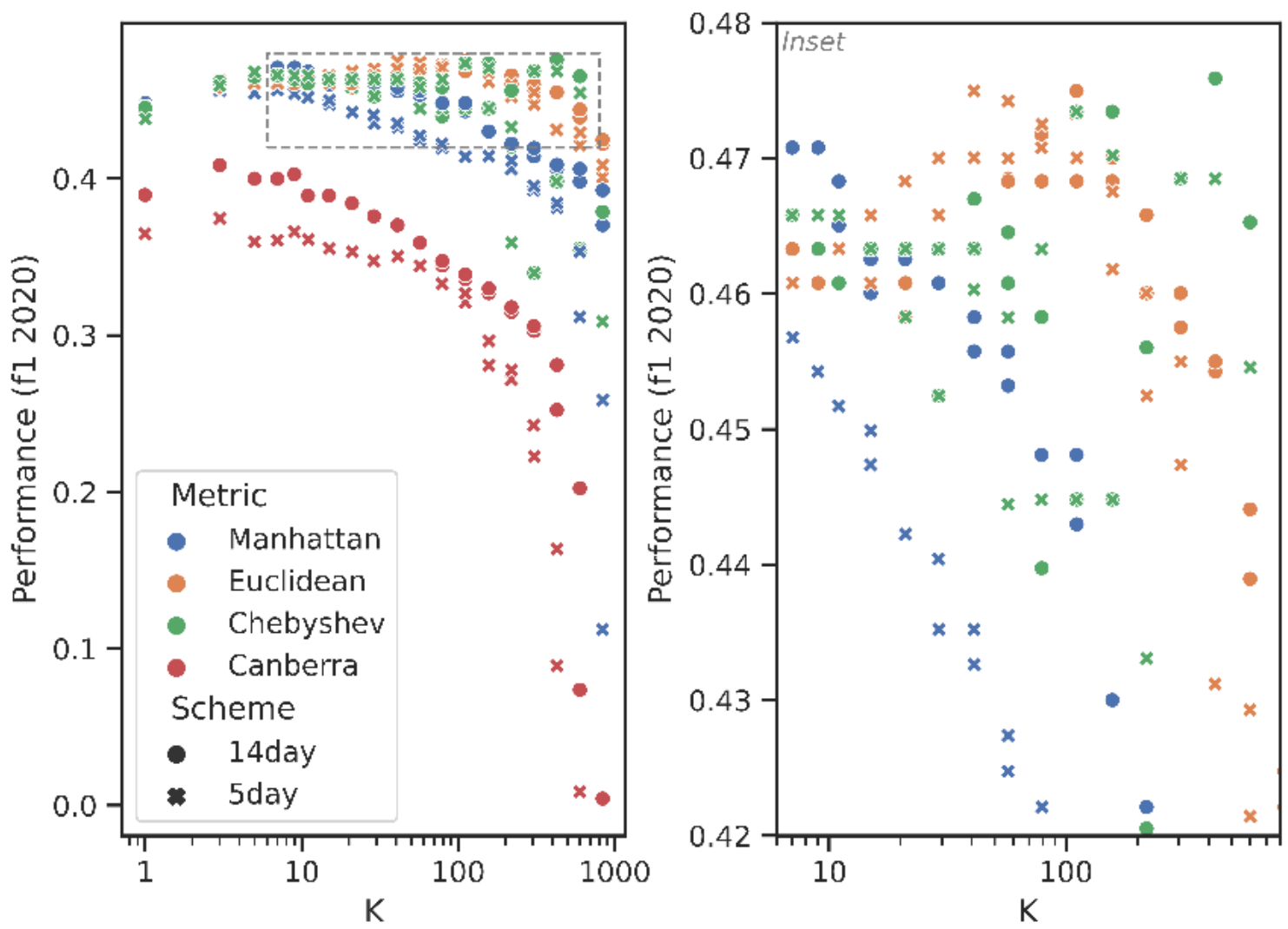

Figure S13. K-nearest Neighbors hyperparameter settings and performance for validation year 2020.

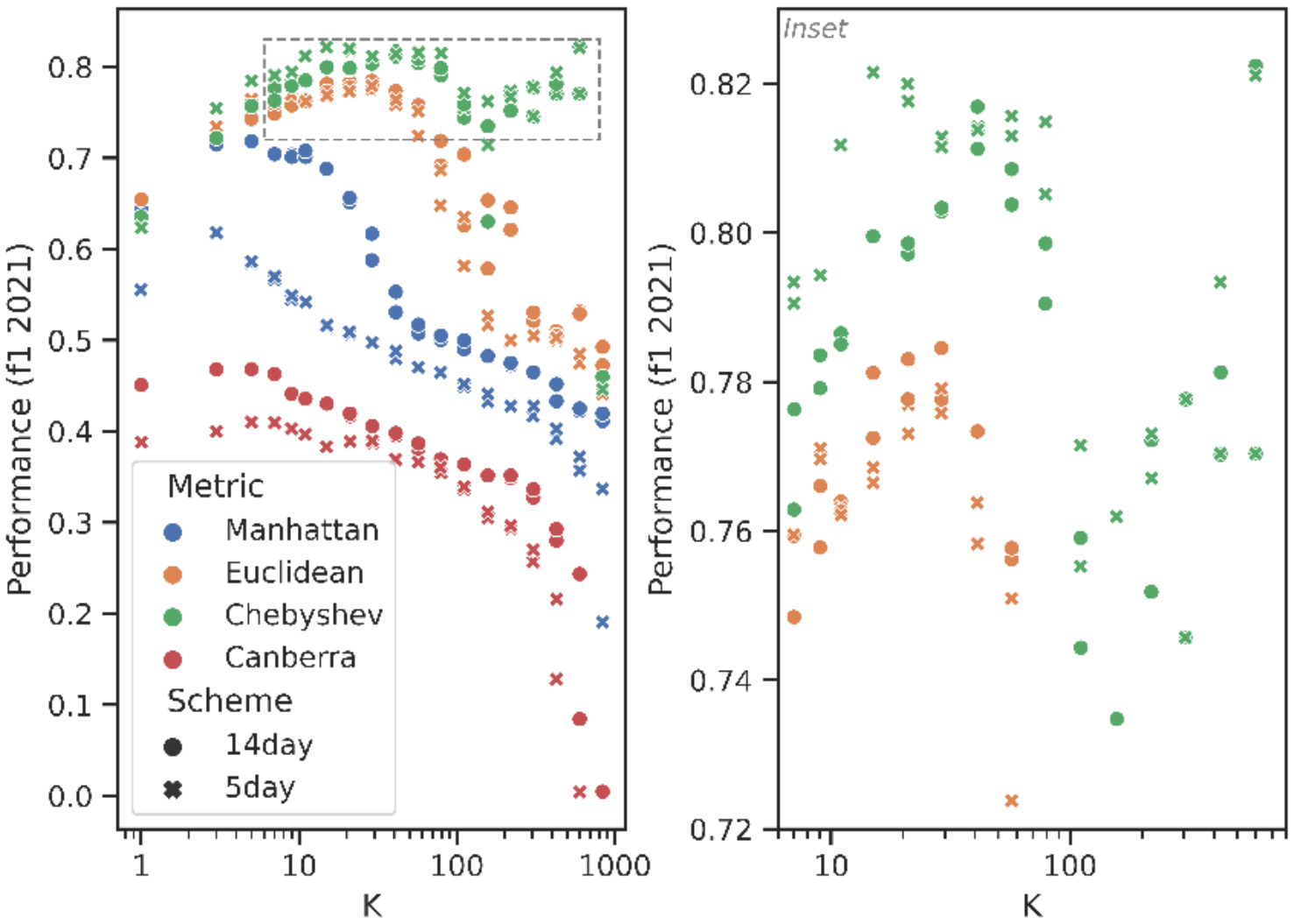

Figure S14. K-nearest Neighbors hyperparameter settings and performance for validation year 2021.

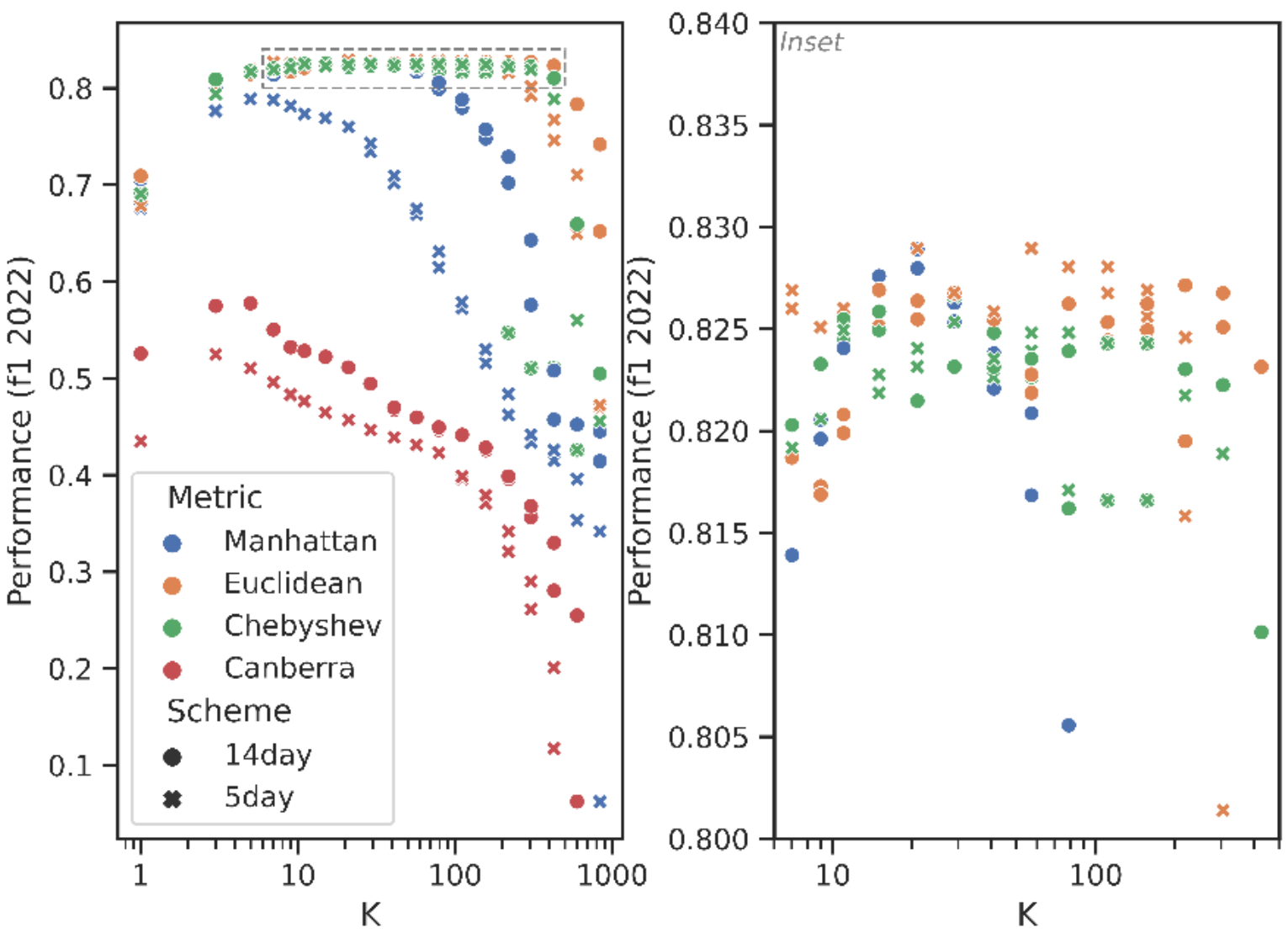

Figure S15. K-nearest Neighbors hyperparameter settings and performance for validation year 2022.